\UseRawInputEncoding

\documentclass[preprint]{vgtc}               

\usepackage{times}                     

\usepackage{tabu}                      
\usepackage{booktabs}                  
\usepackage{lipsum}                    
\usepackage{mwe}                       
\usepackage{mathptmx}                  

\usepackage{times}                     

\usepackage{placeins}
\usepackage{subcaption} 
\usepackage{siunitx} 
\usepackage{amsmath,amssymb,graphicx}
\usepackage{authblk}

\onlineid{0}

\vgtccategory{Research}

\vgtcinsertpkg

\preprinttext{To appear in an IEEE VGTC sponsored conference.}

\title{AR Surgical Navigation With Surface Tracing: Comparing In-Situ
Visualization with Tool-Tracking Guidance for Neurosurgical Applications}

\author[1]{Marc J. Fischer\thanks{e-mail: marcfischer402@gmail.com}}
\author[2]{Jeffrey Potts}
\author[1]{Gabriel Urreola}
\author[3]{Dax Jones}
\author[1]{Paolo Palmisciano}
\author[1]{E. Bradley Strong}
\author[1]{Branden Cord}
\author[1]{Andrew D. Hernandez}
\author[1]{Julia D. Sharma}
\author[1]{E. Brandon Strong\thanks{e-mail: strong@ucdavis.edu }}

\affil[1]{University of California, Davis}
\affil[2]{University of Oklahoma}
\affil[3]{Xironetic}

\teaser{
  \centering
  \includegraphics[width=\linewidth]{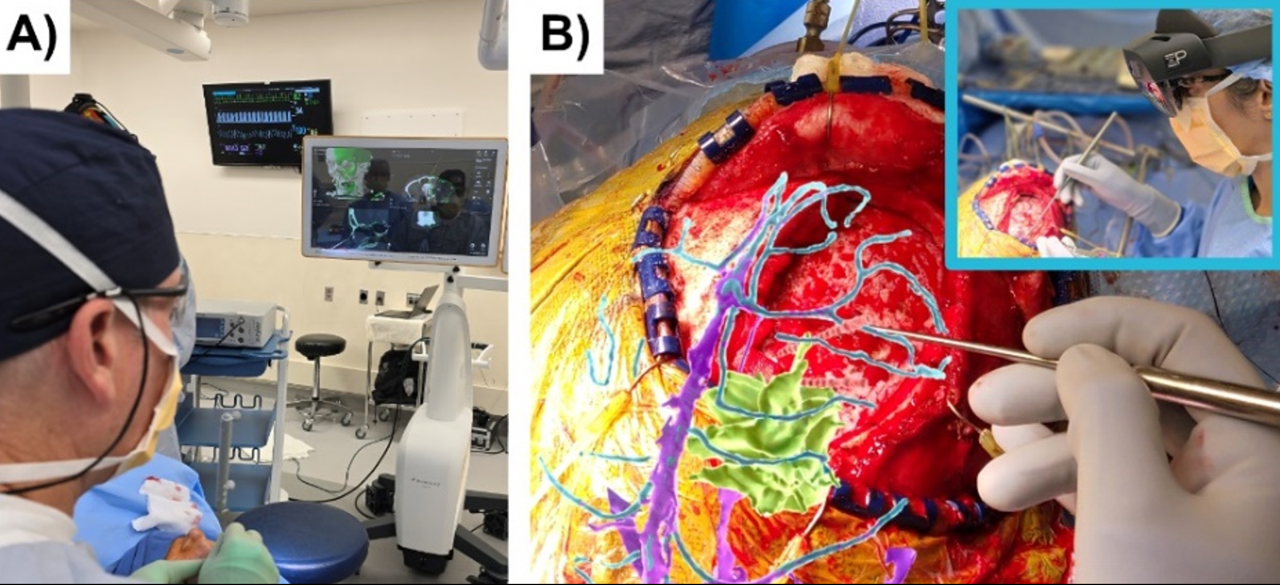}
  \caption{%
    A) Traditional navigation: surgeons look away at a 2D monitor and must mentally map the surgical plan onto the patient in 3D space.  
    B) AR navigation: information is overlaid directly onto patient anatomy in situ, but suffers from depth perception ambiguities.  
    The work presented in this paper aims to address both shortcomings by designing an end-to-end pipeline that leverages state-of-the-art registration and tool-tracking to overlay surgeon critical guidance correctly aligned in their native field of view---without relying on imperfect AR depth cues. We validate this approach in a controlled phantom study as a step toward clinical deployment.%
  }
  \label{fig:teaser}
}

\abstract{
    Augmented Reality (AR) surgical navigation systems are emerging as the next generation of intraoperative surgical guidance, promising to overcome limitations of traditional navigation systems. However, known issues with AR depth perception due to vergence--accommodation conflict and occlusion-handling limitations of the currently commercially available display technology present acute challenges in surgical settings where precision is paramount. This study presents a novel methodology for utilizing AR guidance to register anatomical targets and provide real-time instrument navigation using placement of simulated external ventricular drain catheters on a phantom model as the clinical scenario. The system registers target positions to the patient through a novel surface tracing method and uses real-time infrared tool tracking to aid in catheter placement, relying only on the onboard sensors of the Microsoft HoloLens 2. A group of intended users performed the procedure of simulated insertions under two AR guidance conditions: static in-situ visualization, where planned trajectories are overlaid directly onto the patient anatomy, and real-time tool-tracking guidance, where live feedback of the catheter's pose is provided relative to the plan. Following the insertion tests, computed tomography scans of the phantom models were acquired, allowing for evaluation of insertion accuracy, target deviation, angular error, and depth precision. System Usability Scale surveys assessed user experience and cognitive workload. Tool-tracking guidance improved performance metrics across all accuracy measures and was preferred by users in subjective evaluations. A free copy of this paper and all supplemental materials are available at https://bit.ly/45l89Hq.
}


\begin{document}


\firstsection{Introduction}

\maketitle

\label{sec:introduction}

Many surgical procedures requiring precise placement of instruments along a pre-defined trajectory present significant challenges for surgeons \cite{depthSensorEVD, straighttrack,evdreview,vpshunt}. External ventricular drain (EVD) insertion, ventriculoperitoneal (VP) shunt placement, Kirschner wire (K-wire) placement, burr hole drilling, and many other procedures all depend on the surgeon's ability to accurately translate preoperatively planned trajectories from 2D imaging (e.g., CT, MRI) to the three-dimensional reality of variable patient anatomy.
Traditional approaches rely heavily on anatomical landmarks and the surgeon's spatial understanding, often resulting in multiple attempts and suboptimal outcomes.

\subsection{Neurosurgery Clinical Context}

When swelling, bleeding, or tumour growth raises pressure inside the skull, neurosurgeons insert a catheter into the fluid-filled cavities (ventricles) of the brain to relieve the pressure \cite{Muralidharan2015,evdorvsicu}. In emergencies, the catheter exits the scalp and drains into a calibrated chamber, allowing the team to monitor and adjust pressure in real time\cite{Savage2021,Malone2019}.
The standard ``free-hand'' method typically used in this scenario is quick and requires no special hardware. With the patient lying supine and the head slightly raised, the surgeon makes a C-shaped incision and drills a burr hole about 10 cm behind the bridge of the nose and 3 cm to the right side, then advances the catheter roughly 7 cm toward the midline until clear fluid flows \cite{Huyette2008,Muralidharan2015,Fowler2025}. Although this landmark-guided technique is life-saving, it misses the ventricle in 10--40\% of cases, risking bleeding, infection, and early shunt blockage \cite{evdreview,kakarla2008safety,sarrafzadeh2014guided,Kamenova2018,Ajmera2019}.

Image-guided navigation using infrared or electromagnetic tracking is commonplace in numerous neurosurgical interventions and has been shown to significantly improve catheter-placement accuracy \cite{AlAzri2017,Jakola2014,Kamenova2018,Ajmera2019,Kullmann2018}; however, it remains uncommon for EVD procedures because the equipment often requires a lengthy setup, carries a high price tag, and can be cumbersome to use \cite{Patil2015}. In a nationwide survey, more than half of participating neurosurgeons said they would employ an image-guided system that guaranteed accurate catheter placement, provided the entire setup could be completed in under ten minutes \cite{Patil2015,ONeill2008}.

\subsection{Augmented Reality}

Augmented reality (AR) shows promise in delivering the same benefits as traditional image guidance, while overcoming its cost and setup time drawbacks. A self-contained HoloLens 2 headset costs roughly \$3,500 vs.\ several hundreds of thousands for a traditional navigation system \cite{Asadi2025}, is lightweight, and can be set up quickly. Additionally, because its sensors are less than a meter from the operative field on the surgeon's head, AR can avoid the line-of-sight limitations of external tracking cameras. By overlaying navigation guidance directly on the patient, AR can also eliminate the need to glance back and forth between a remote monitor and the surgical site, reducing cognitive load and streamlining the surgeon's workflow.

An AR navigation pipeline can be divided into two core stages: registration---establishing the spatial relationship between pre-operative images and the patient---and navigation---maintaining that relationship in real time while presenting relevant guidance. A broad range of techniques has been investigated for each stage.

\subsubsection{Registration}

Many studies rely on an external tracking system, but this introduces additional sources of error: the tracker must be calibrated to the headset, line-of-sight can be lost, and overall set-up time and cost increase.

Other work instead exploits the headset's own sensors, thereby avoiding these drawbacks. CT-based registration, surface tracing, and surface scanning are established image-based guidance methods used by Brainlab and Medtronic that can deliver a much higher degree of accuracy than purely perception-based manual alignment\cite{VentroAR, Schneider2021Augmented} or landmark-only registration without surface tracing \cite{ChristophKabsch,AzimiBedside}. Several works have translated these traditional methods to run on the HoloLens 2, leveraging its onboard sensors.
For example, Zhang et al.\ \cite{straighttrack} used CT imaging for patient registration in K-wire placement, employing a reflective-marker array tracked by the headset's time-of-flight sensor to transform between the CT-image coordinate system and patient space. Although this approach is highly accurate, it is impractical for routine use because of increased set-up time, radiation exposure, and cost.

Liebmann et al.\ developed an AR orthopaedic-navigation prototype for the first-generation HoloLens that localises a stylus by detecting ArUco fiducials. Stylus-tip positions are recorded only while the user depresses a Bluetooth trigger, producing dense, intentional surface point clouds ($\approx 1{,}983$ points on average) that are subsequently registered to the pre-operative model via ICP. A key limitation is that marker detection in visible light is less precise \cite{martin2023sttar} and more sensitive to lighting than infrared tracking; moreover, the ageing HoloLens 1 hardware lacks the improved capabilities of the HoloLens 2. Van Gestel et al.\ \cite{vanGestel2023NeuroAR} tackle these limitations on the HoloLens 2 with an infrared-based landmarking-plus-surface-tracing workflow, but report a tool-tracking accuracy of 0.78 $\pm$ 0.74 mm without describing their measurement procedure. This omission prevents direct comparison with the 0.1 mm lateral, 0.5 mm depth, and $< 1^\circ$ rotational errors reported by Gomez et al.\ \cite{martin2023sttar}. Moreover, their proprietary algorithm is also used to track the reference array used to align the phantom CT to HoloLens space as a registration ground truth, yet the array's own angular error---and the abbe error amplification introduced by positioning it some distance from the phantom---remain unquantified. Because the underlying tracking method is neither described nor independently validated, the overall registration accuracy is difficult to judge and the results are hard to replicate.

Li et al.\ \cite{depthSensorEVD} placed infrared-reflective flat fiducials on the patient's skin to correct HoloLens 2 depth-sensor data and accurately reconstruct the patient's head surface. They reported a substantial raw depth error on human skin (7.580 $\pm$ 1.488 mm) and developed a patient-specific depth-correction model that reduced this error by more than 85\%. The reconstructed head surface was then registered to pre-operative images using iterative closest point (ICP) registration. In tests on a real sheep head, the surface reconstruction achieved sub-millimetre accuracy, with a reconstruction error of 0.79 mm \cite{depthSensorEVD}.
This approach appears very promising. One drawback is that it is less familiar to surgeons than stylus tracing and requires more extensive validation in humans, as it is a relatively novel method that has been tested on only 10 subjects. It also provides a smaller registration area on patients with hair, whereas a stylus can usually be manoeuvred between hair strands. An advantage, however, is that the method is contactless; therefore, it avoids the skin-compression errors introduced when a stylus presses on the skin.
A major bottleneck is that the current implementation takes almost ten minutes to perform calibration and registration, with model-parameter identification alone requiring 397.968 $\pm$ 194.462 s---much longer than using surface tracing \cite{depthSensorEVD}.

\subsubsection{Navigation}
AR navigation techniques can be grouped into two visual--interaction paradigms:  
(i) \emph{static in-situ overlays}, in which the anatomy and planned trajectory are rendered once on the patient and remain unchanged during the procedure, and  
(ii) \emph{tool-referenced feedback}, in which the display continuously updates to reflect the live six-degree-of-freedom pose of the instrument.

Many systems adopt the simplest metaphor---a holographic line connecting the skin entry point to the intracranial target. Li et al.\ \cite{depthSensorEVD} and Azimi et al.\ \cite{AzimiBedside} exemplify this approach. Projecting the plan directly on the patient helps surgeons localise the burr hole, but because depth and angulation must be inferred from 3-D objects rendered with imperfect depth cues, overshoot and trajectory drift remain common. For example, in their simulated EVD procedure Li et al.\ achieved an entry-point accuracy of 2.09 $\pm$ 0.16 mm and an orientation accuracy of 2.97 $\pm$ 0.91$^\circ$.

Another approach is to render only the tracked catheter as an overlay so that it can be visualized inside the patient. Sun et al.\ render a live virtual twin of the catheter---whose five-degree-of-freedom (5-DoF) pose is supplied by their external colour-band tracker---to make the catheter visible inside the skull \cite{5DoFTracking}. However, vergence--accommodation conflict, occlusion-handling challenges in optical see-through (OST) displays, and the absence of a registered CT volume and surgical plan against which to reference the tool limit the usefulness of this approach.

Benmahdjoub et al.\ \cite{Benmahdjoub2021} combines and studies these two approaches: they render the planned trajectory and then align the tool to it, with and without a tracked virtual overlay. They found that the virtual overlay on the tracked tool significantly improved alignment accuracy and showed that it is easier to align two virtual objects overlaid on a real object than to align a real object with a virtual one.

Liebmann et al.\ \cite{liebmann2020registrationeasystandalone} provide guidance with such tool-tracked navigation via a holographic triangular display overlaying the planned and actual tool trajectories. In phantom experiments (n = 2) they achieved a registration RMSE of 1.62 mm, a mean entry-point error of 2.77 $\pm$ 1.46 mm, and a trajectory-angle error of 3.38 $\pm$ 1.73$^\circ$, with digitisation taking about 125 $\pm$ 27 s using a different registration approach than Li et al.\ \cite{depthSensorEVD}.

Ventricular catheter placement, however, is fundamentally an egocentric real-to-real task: the surgeon must steer a physical catheter through tissue along a pre-planned trajectory, controlling the entry point, angle, and depth. A 3-D holographic overlay---prone to depth-perception ambiguities---is not strictly necessary. Instead, mapping essential cues such as distance to the entry point, angle, and depth to target into a 2-D overlay within the surgeon's egocentric plane, while preserving a clear view of the target as done in \cite{straighttrack}, may be more effective, guiding the catheter step by step rather than attempting to align it all at once as studied in Benmahdjoub et al.\ \cite{Benmahdjoub2021} and occluding the target area and tool\cite{liebmann2020registrationeasystandalone}.

Encoding some cues e.g., depth in color\cite{VentroAR} or using sound \cite{Schuetz04072023} has also been proposed to increase the usability of such systems.

\subsection{Contribution}
We present and validate a reproducible, state-of-the-art registration workflow for the HoloLens 2 that combines landmark digitisation with surface tracing using an optically tracked stylus powered by a non-proprietary, validated tool-tracking algorithm \cite{martin2023sttar}. Although conceptually identical to the workflows of conventional surgical-navigation systems, its familiarity to surgeons and completion time of less than two minutes make it markedly more practical for external ventricular drain (EVD) placement than the approximately 10-minute surface-scanning method proposed by Li et al.\,\cite{depthSensorEVD, ONeill2008}.

To streamline tracing, we introduce an automatic outlier-rejection strategy that discards stylus samples collected off the anatomical surface, thereby obviating the need for manual triggers such as clickers\cite{liebmann2020registrationeasystandalone}, foot pedals or other sensing modalities like heat or pressure on stylus tip allowing the user to effectively surface trace to a target with a passive stylus.

Whereas Benmahdjoub et al.\,\cite{Benmahdjoub2021} compared the alignment of two virtual 3-D objects against aligning a virtual object to a real object, we compare to a navigation user interface (UI) that provides real-time quantitative feedback without relying on depth cues, which are poorly reproduced by the commercially available display technology, as proposed by several groups in previous studies \cite{straighttrack, liebmann2020registrationeasystandalone}.

Implementing this registration workflow for catheter insertion in a model study, we present results showing how real-time trajectory guidance affects target accuracy compared with static AR trajectory overlays, thereby informing continued progress towards a broadly applicable AR navigation system for EVD placement. Finally, this workflow could find broad utility across numerous surgical disciplines and may further accelerate the adoption of this promising technology.

\begin{figure*}[t]
  \centering
  \subcaptionbox{Experimental setup\label{fig:exp_setup}}
    {\includegraphics[height=3.8cm]{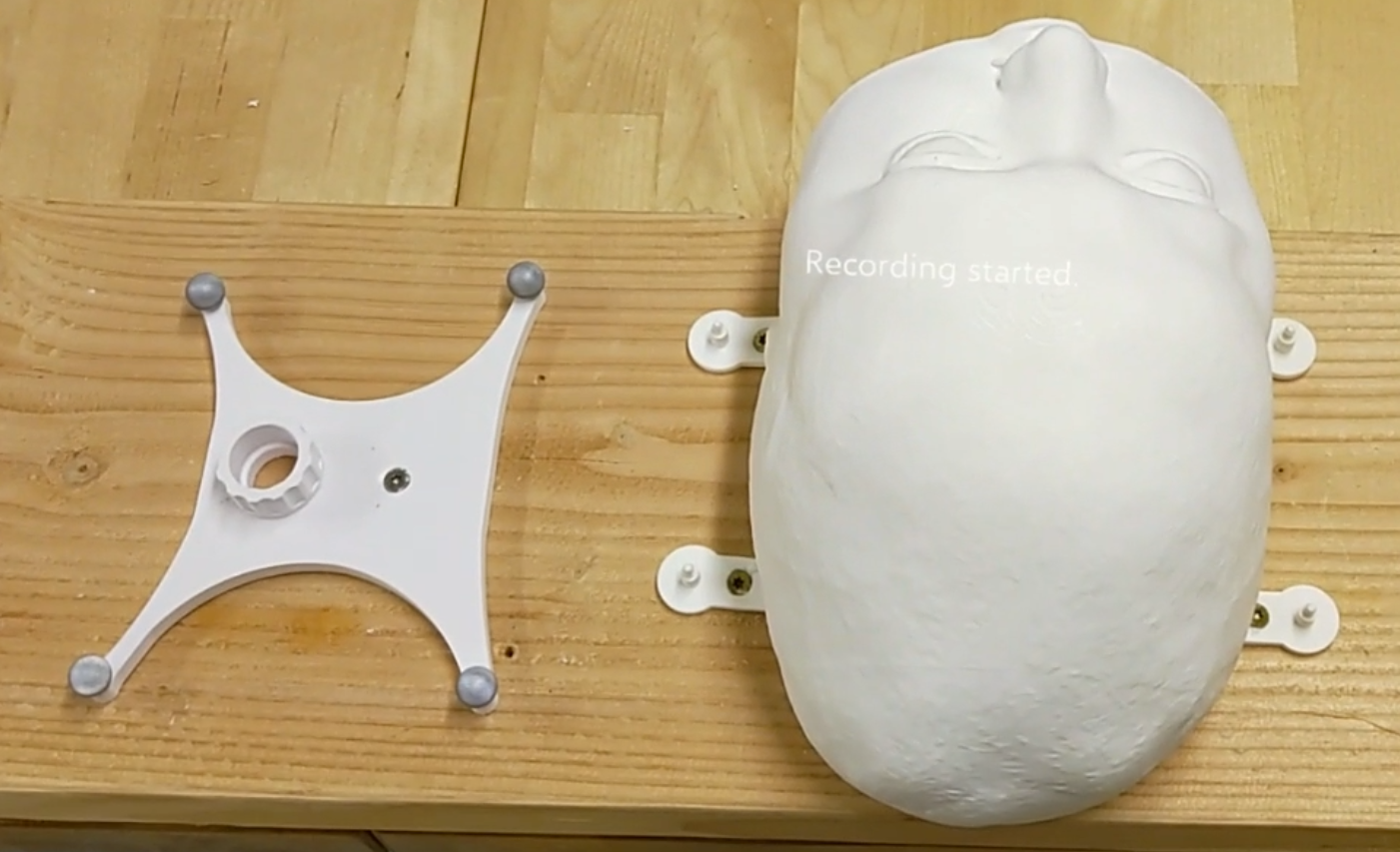}}\hspace{0.6em}
  \subcaptionbox{Rough registration \\ (landmark-based)\label{fig:landmarks}}
    {\includegraphics[height=3.8cm]{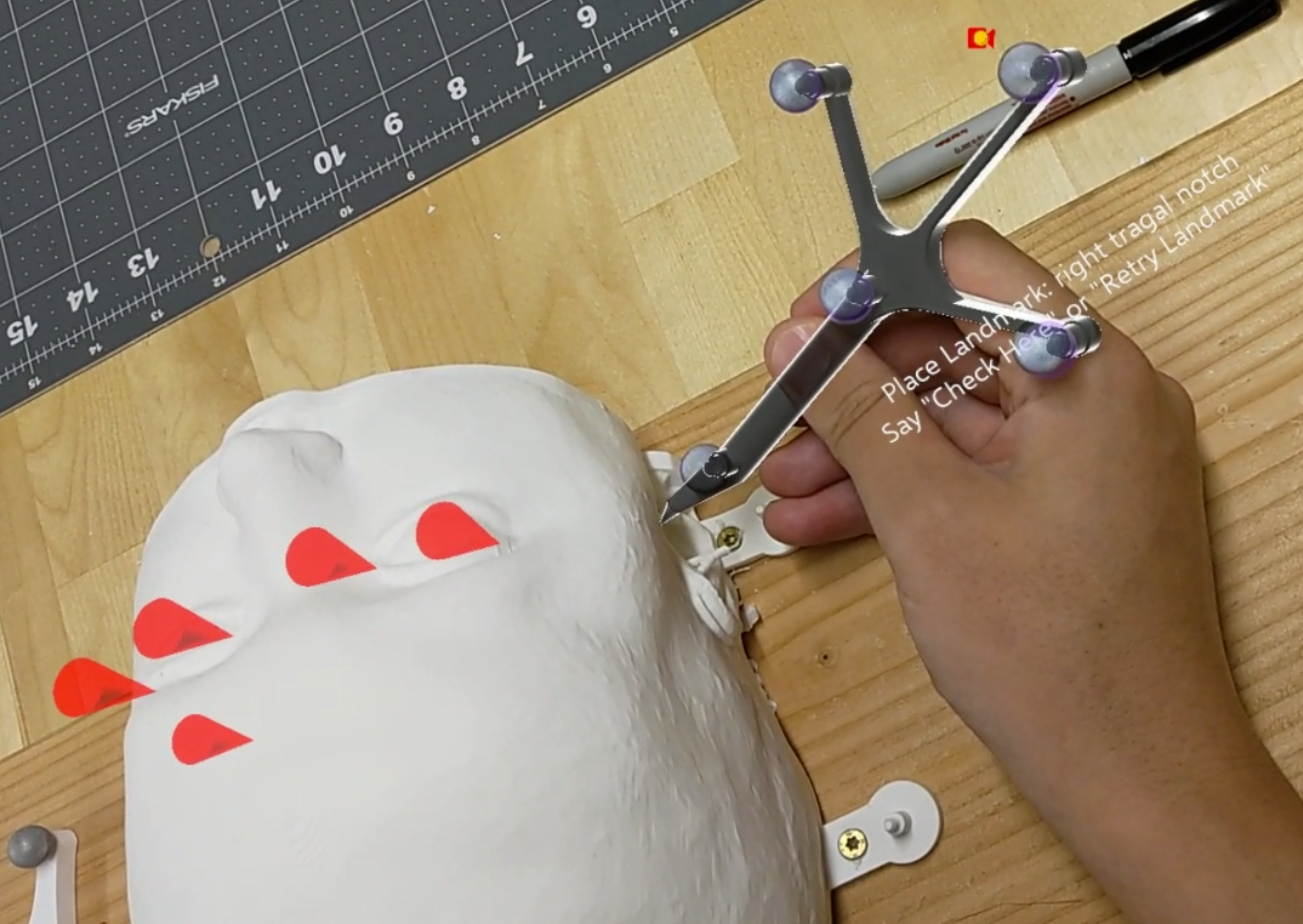}}\hspace{0.6em}
  \subcaptionbox{Fine registration \\ (surface tracing)\label{fig:tracing}}
    {\includegraphics[height=3.8cm]{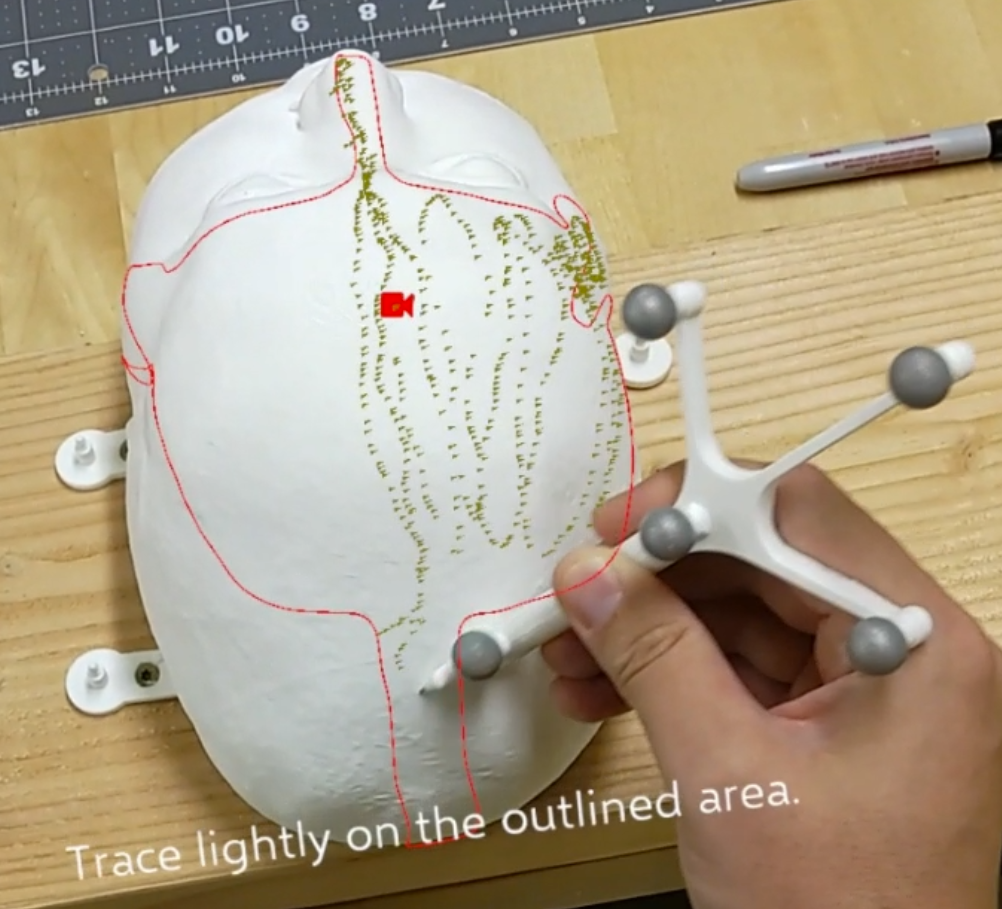}}

  \vspace{0.9em}

  \subcaptionbox{Marking burr hole \\ (in-situ overlay)\label{fig:marking_is}}
    {\includegraphics[height=3.8cm]{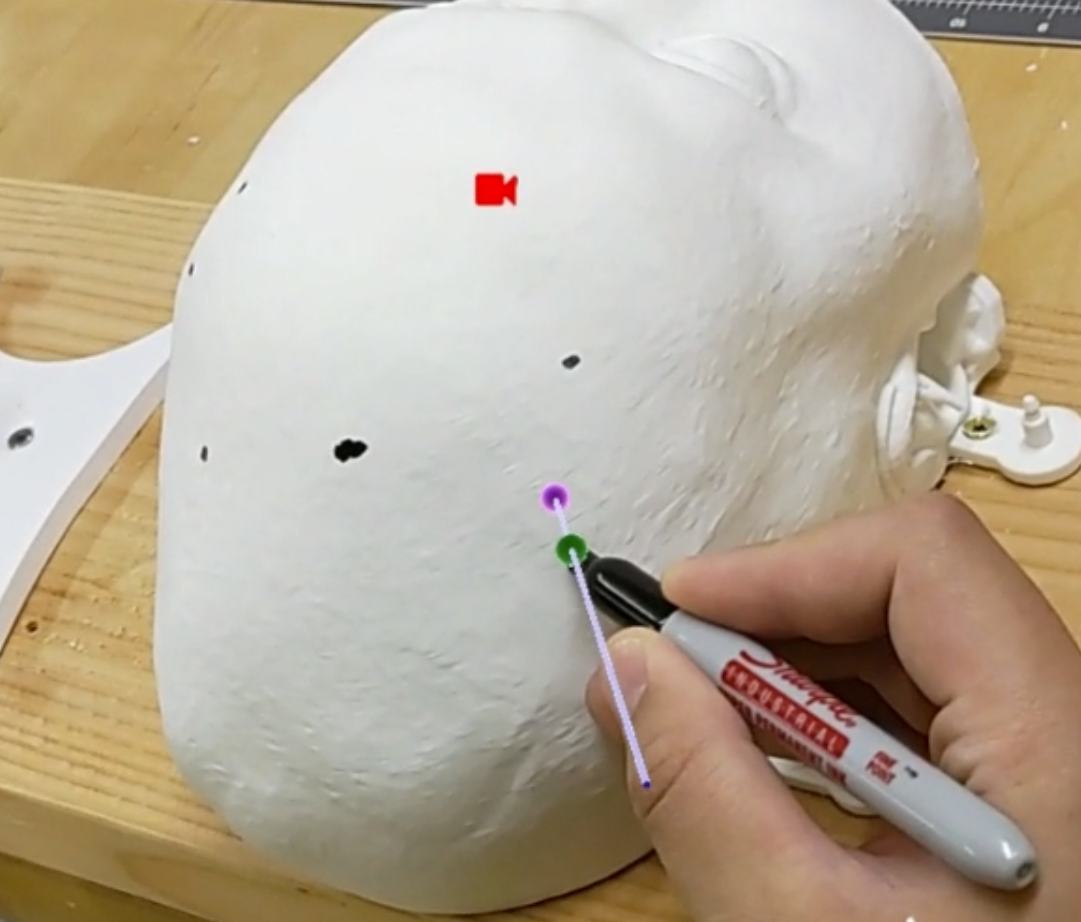}}\hspace{0.6em}
  \subcaptionbox{Marking burr hole \\ (tool-tracked)\label{fig:marking_tt}}
    {\includegraphics[height=3.8cm]{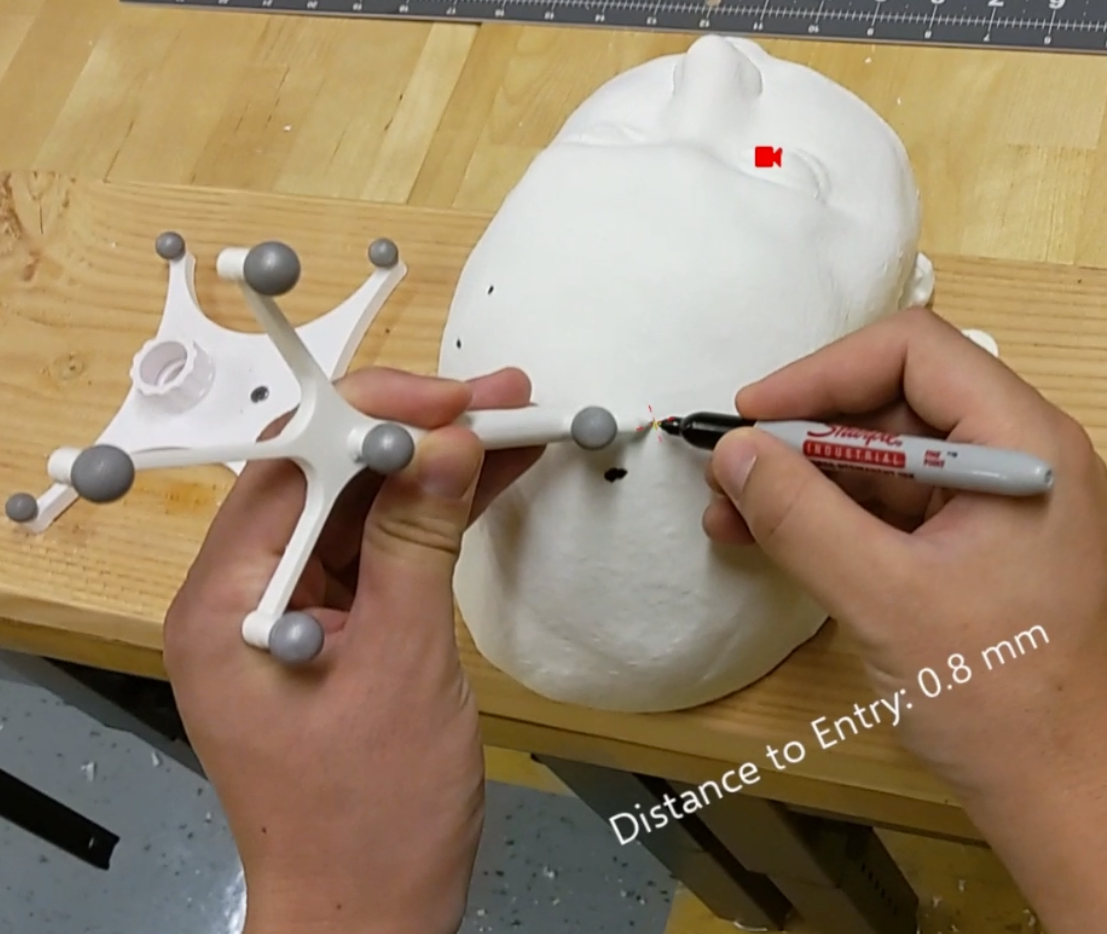}}\hspace{0.6em}
  \subcaptionbox{Drilling burr hole\label{fig:drilling}}
    {\includegraphics[height=3.8cm]{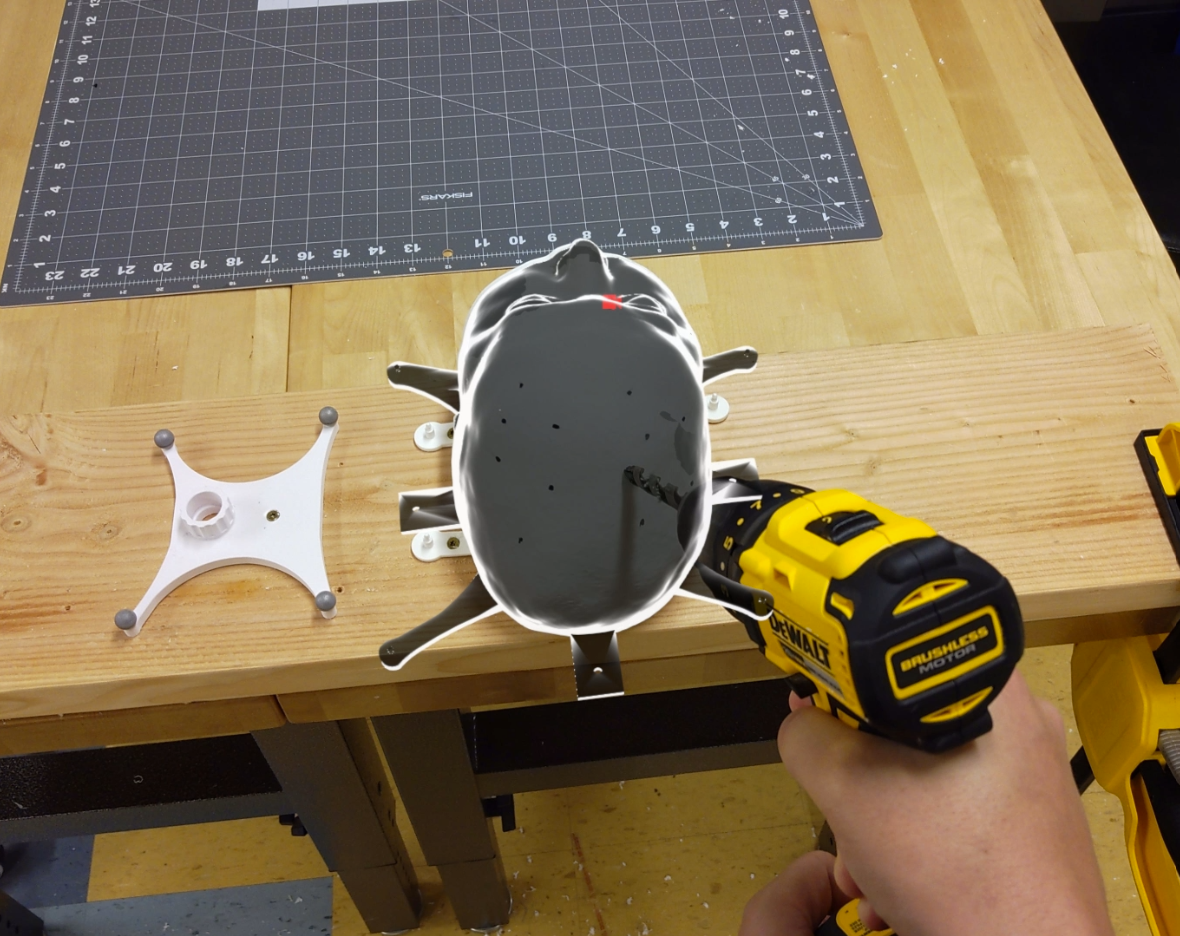}}

  \vspace{0.9em}

  \subcaptionbox{Catheter insertion \\ (in-situ view)\label{fig:insitu}}
    {\includegraphics[height=3.8cm]{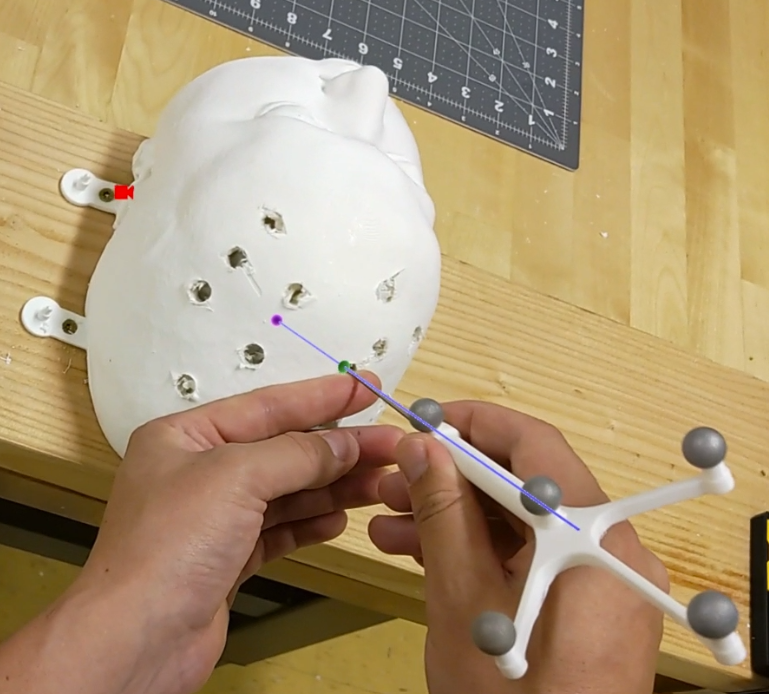}}\hspace{0.6em}
  \subcaptionbox{Tool-tracked insertion \\ (view 1 - off trajectory)\label{fig:tool1}}
    {\includegraphics[height=3.8cm]{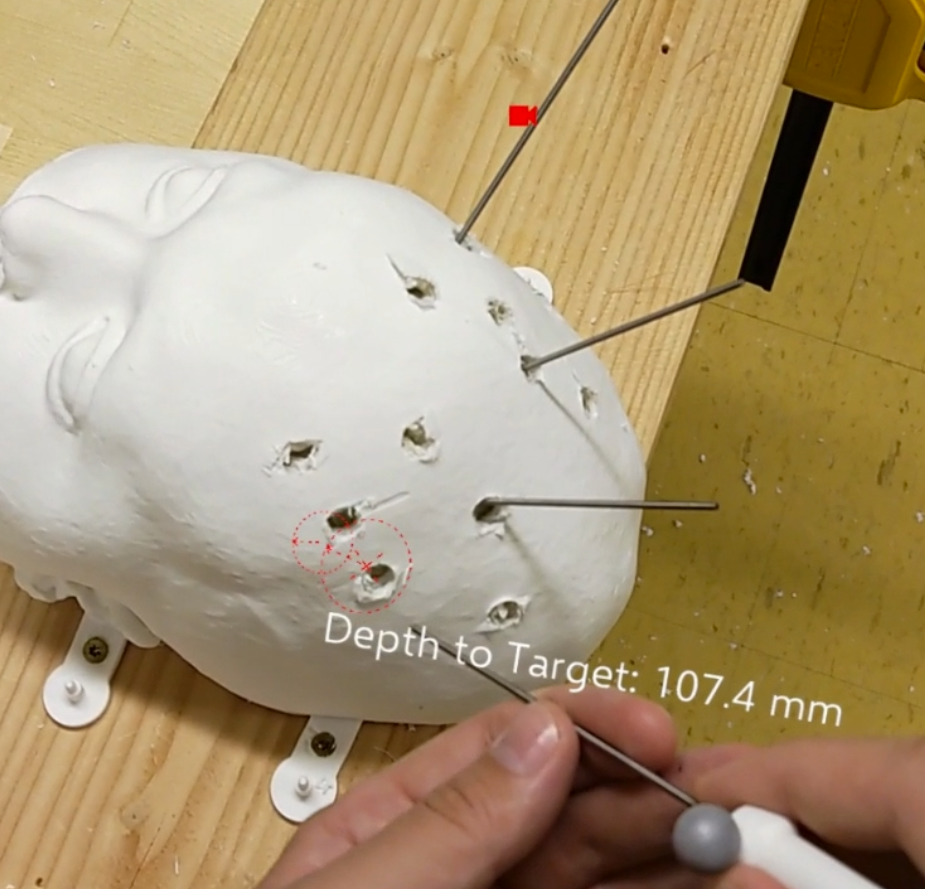}}\hspace{0.6em}
  \subcaptionbox{Tool-tracked insertion \\ (view 2 - on trajectory)\label{fig:tool2}}
    {\includegraphics[height=3.8cm]{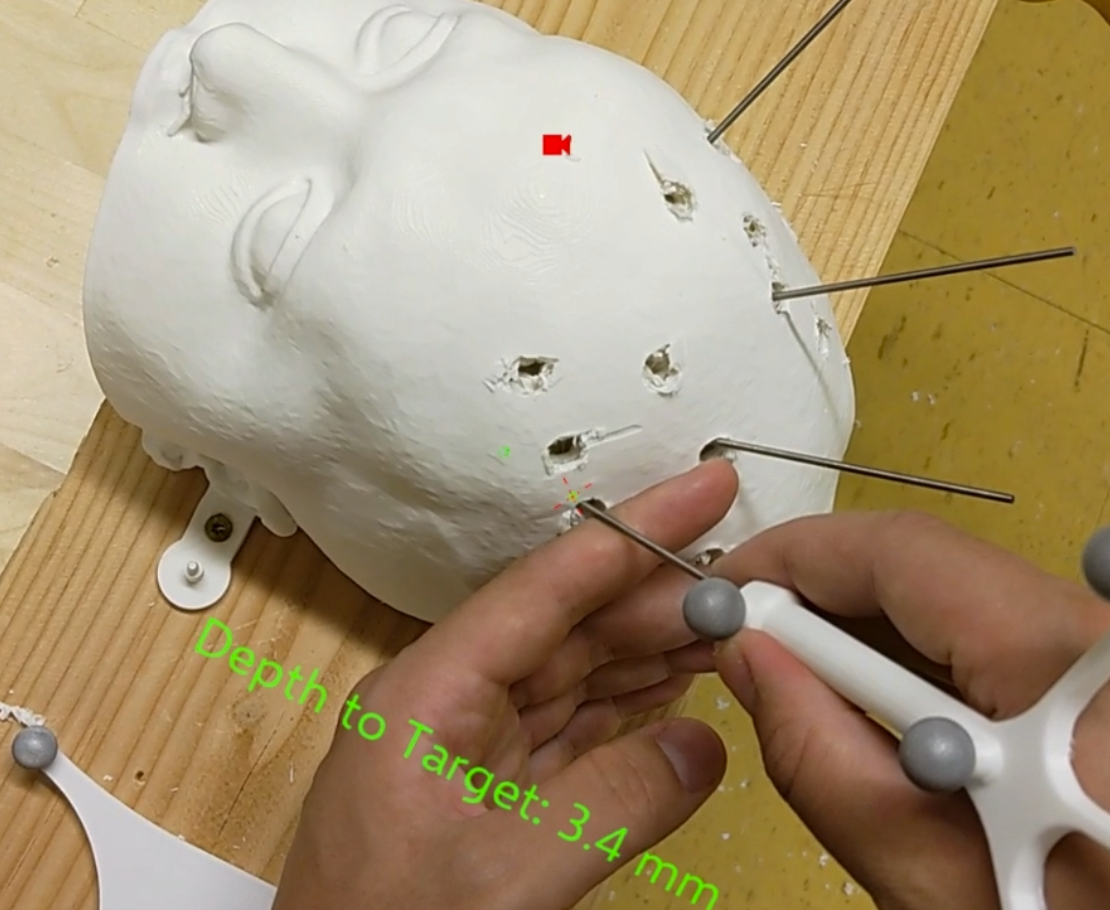}}

  \caption{Augmented-reality navigation workflow for neurosurgical catheter placement. After landmark and surface-tracing registration (\subref{fig:exp_setup}--\subref{fig:tracing}), burr-hole marking can be performed either with static in-situ overlays (\subref{fig:marking_is}) or with real-time tool-tracking assistance (\subref{fig:marking_tt}). Subsequent drilling and catheter insertion similarly benefit from either visualisation mode (\subref{fig:drilling}--\subref{fig:tool2}). To see the full experimental workflow, please refer to the video in the supplemental materials.}
  \label{fig:ar_workflow}
\end{figure*}

\section{Methods}

\subsection{System Workflow}

We developed a system that allows the surgeon to register preoperative imaging data as 3D segmented models as well as planned trajectories to the patient and then receive guidance based on how a tracked catheter is positioned relative to the plan. 

\subsection{Registration}
For registration, we track the pose of a custom, additively manufactured stylus derived from a commercially-available surgical navigation stylus design. This stylus is outfitted with retroreflective spheres for infrared tracking (OptiTrack, Oregon, USA) using the HoloLens 2's Articulated Hand Tracking (AHAT) sensor accessible through Research Mode. We build upon prior work on infrared array tracking with the HoloLens 2 \cite{martin2023sttar, keller2023hl2irtracking}. The tip of our custom stylus is coincident with the mid plane of the retroreflective spheres and as close as possible to the tool center to minimize tip tracking error \cite{west}. The nominal tip position of our custom stylus is known from computer-aided design software and is verified by caliper measurements, thereby allowing for the tip position to be calculated from each successful detection of the infrared array by the AHAT sensor. 

The user initializes the registration sequence using a predefined voice command to the HoloLens. The user is then guided to sequentially indicate six landmarks on the phantom model using the stylus (left and right intertragal notch, lateral canthi, and medial canthi). A point-to-point rigid transformation is applied to align the virtual anatomy model to the phantom \cite{Umeyama}, providing an initial alignment that can be refined with our surface tracing method. The user is then instructed to trace a predefined area of the phantom while keeping the stylus in contact with the model surface. The stylus tracing process generates a source point cloud that can be aligned to the target point cloud sampled from the virtual anatomy model for refinement of the initial registration obtained from the landmark registration step. We detail the key elements of our registration algorithm in the following section.

\subsection{Surface Tracing Registration Algorithm}
The landmark registration process for initial registration follows methods previously reported which estimate the virtual anatomy model pose by minimizing the mean-squared error between indicated and virtual landmarks using an analytical solution based on singular value decomposition \cite{Umeyama}. Following this initial registration, we have:

\begin{itemize}
    \item A dense target point cloud $P_{s} = \{\mathbf{x}_i\}$ with normals $\{\mathbf{n}_i\}$.
    \item Incoming traced points $Q = \{\mathbf{q}_j\}$ collected by the tracked stylus tip.
    \item A KD-tree on $P_s$ for efficient nearest-neighbor queries.
\end{itemize}

For each surface point $\mathbf{x}_i \in P_s$, we identify a subset of traced points $Q_i \subset Q$ that are associated with $\mathbf{x}_i$ via the KD-tree nearest-neighbor search. We compute a robust representative trace point $\mathbf{t}_i$, reducing the influence of noise, jitter and outliers in the raw traced data as explained in the following section.

\subsubsection{Inlier Filtering}
Each incoming point $\mathbf{q}_j$ is assigned to its nearest neighbor $\mathbf{x}_{i^*} \in P_s$, where:
\[
i^* = \arg\min_i \|\mathbf{q}_j - \mathbf{x}_i\|.
\]
We apply an inlier filtering scheme where if $\|\mathbf{q}_j - \mathbf{x}_{i^*}\| \leq 10\text{ mm}$, then $\mathbf{q}_j$ is considered an inlier and is included in the correspondence set. We choose 10 mm as our threshold to accommodate variation from the landmark registration process, which in our prior testing has typically produced fiducial registration error under 10 mm by trained users (unreported results).

\subsubsection{Outlier Removal and State Handling}
We define the system to be in an in-bounds state at step $j$ if:

\begin{itemize}
    \item The incoming point $\mathbf{q}_j$ satisfies the inlier condition:
    \[
    \| \mathbf{q}_j - \mathbf{x}_{i^*} \| \leq 10~\text{mm},
    \]
    where $\mathbf{x}_{i^*}$ is the nearest point in the target point cloud.
    
    \item No previous outlier has triggered a transition to an out-of-bounds state.
\end{itemize}

An out-of-bounds state occurs when the inlier condition fails while the system is in-bounds. We define a re-entry point $\mathbf{r}$ for state transitions. The system evolves according to the following logic:
\begin{itemize}
    \item In-bounds, inlier: If the system is in-bounds and $\mathbf{q}_j$ is an inlier, link it to its corresponding target point $\mathbf{x}_{i^*}$.
    
    \item In-bounds, outlier: If $\mathbf{q}_j$ exceeds the inlier threshold, remove any recently added points within 10 mm of $\mathbf{q}_j$, transition to the out-of-bounds state, and clear $\mathbf{r}$.
    
    \item Out-of-bounds, candidate re-entry: If the system is out-of-bounds, $\mathbf{q}_j$ is an inlier, and $\mathbf{r}$ is unset, set $\mathbf{r} := \mathbf{q}_j$.
    
    \item Re-entry condition met: If $\mathbf{r}$ is set and $\| \mathbf{q}_j - \mathbf{r} \| > 10~\text{mm}$, resume tracing by transitioning to the in-bounds state and linking $\mathbf{q}_j$ to $\mathbf{x}_{i^*}$.
\end{itemize}

\subsubsection{Normal-Based Projection Filtering}
We apply a projection-based filter to eliminate traced points that deviate too far from the surface plane defined by the normal $\mathbf{n}_i$. For each index $i$ in the target point cloud $P_s$, given $\{\mathbf{q}_j\}$:
\[
p_j = \mathbf{q}_j \cdot \mathbf{n}_i.
\]
Let $p_{\min} = \min_j p_j$. Discard any point $\mathbf{q}_j$ where $(p_j - p_{\min}) > 3\text{ mm}$.

From the remaining inliers, compute an estimate of the local traced point:
\[
\mathbf{t}_i = \text{median}\{\mathbf{q}_{j}\ \text{remaining}\}.
\]

\subsubsection{Alignment}
\label{subsubsec:alignment}

Let $P$ be the sparse point cloud traced on the patient (phantom) and $Q$ a dense point cloud of the pre-operative anatomy model.  
We seek a rigid transform $T\colon P \!\to\! Q$ that maximises geometric agreement.  
Starting from the coarse landmark-based registration, we score this initial alignment using the criteria detailed below and then refine it iteratively using a \emph{multiscale iterative closest point (ICP)} algorithm.  

We denote the landmark-based seed by $T_0$ and keep a running \emph{current best} transform $T$ (initially $T=T_0$).  
Each ICP call estimates an \emph{incremental} rigid transform $\Delta T$; composing it with the current best gives a candidate refinement $T_{\text{new}} = \Delta T \circ T$.  
Starting on heavily down-sampled clouds, the algorithm estimates a global pose, then iteratively tightens the correspondence radius on progressively denser clouds.  
This coarse-to-fine strategy is less sensitive to initialization and local minima than single-resolution (``vanilla'') ICP.

At each scale, we voxel--down-sample both the sparse traced cloud and the dense model with a voxel size
$\tau_i$ drawn from a logarithmically spaced series of 15 levels spanning
\SI{10}{\milli\metre} to \SI{0.1}{\milli\metre} (hand-tuned, based on unreported results):
\[
\{\tau_1,\dots,\tau_{15}\} \;=\;
\operatorname{logspace}\!\Bigl(
    \log_{10}\bigl(\SI{10}{\milli\metre}\bigr),
    \log_{10}\bigl(\SI{0.1}{\milli\metre}\bigr),\,
    15
\Bigr).
\]

Point-to-point ICP from Open3D\cite{zhou2018open3d} starts from the current best transform $T$, runs for at most 200 iterations, and stops when relative fitness and RMSE change by less than $10^{-6}$.  
The maximum correspondence distance is proportional to $\tau_i$.

\paragraph{Scoring}  
For any candidate transform $T$ we compute
\[
\text{Fitness}
    = \frac{N_{\mathrm{close}}}{N_{\mathrm{traced}}},
\qquad
\text{RMSE}
    = \sqrt{\frac{1}{N_{\mathrm{close}}}
            \sum_{i=1}^{N_{\mathrm{close}}}
            \bigl\lVert \mathbf{p}_i^{T} - \mathbf{q}_i \bigr\rVert^{2}},
\]
where $N_{\mathrm{traced}}$ is the number of traced points and $N_{\mathrm{close}}$ counts those whose nearest neighbour on $Q$ lies within \SI{5}{\milli\meter}.  
The score
\[
S = 0.4\,\text{Fitness}
    + 0.6\!\left(1 - \frac{\text{RMSE}}{2 \times \SI{5}{\milli\meter}}\right)
\]
rewards dense correspondences and low residual error (hand-tuned
based on unreported results).  
Alignments with $\text{Fitness} < 0.7$ or $\text{RMSE} > \SI{5}{\milli\meter}$ are rejected $(S = 0)$.

If the score $S$ of the refined transform exceeds the current best, we update $T \leftarrow T_{\text{new}}$; otherwise the previous best is retained.  
Because every ICP call starts from the best-so-far $T$, $P$ is always refined in its presently aligned frame.  
After all scales, the inverse of the highest-scoring transform aligns the dense anatomy model $Q$ to the traced cloud $P$.

\subsubsection{Guidance}\label{subsubsec:guidance}
Following alignment, a user interface can be activated for real-time trajectory guidance of the simulated catheter, which is tracked by an infrared array that can be easily detached upon placement into the phantom model. In this study, users perform simulated catheter insertions with and without this real-time guidance, but in both cases with AR overlays for reference. We deem the scenario of active AR guidance of catheter position as the \textit{tool tracking} approach and the static AR trajectory overlays as \textit{in situ} (IS) visualization.

In situ visualization shows an AR overlay of the entry point for drilling or catheter placement as a 6 mm-diameter disc and the trajectory as a 1 mm diameter cylinder spanning from the target to the entry point and protruding 10 cm out of the skin surface. The target point is visualized as a 4 mm sphere. These static overlays can be visualized with an outline rendering of the patient's skin. 

For the tool tracking approach, we track the tip position of the catheter and the tool trajectory of the catheter. For the initial marking of where to drill the burr hole, we compute a plane on the planned entry point with the planned trajectory as its normal. We place a plane perpendicular to the normal of the planned trajectory of each entry point and find the intersection points of the catheter trajectory and the planes. We build upon the approach of Zhang et.\ al by using circles in each plane indicating the distance between the intersection of the actual trajectory and the planned entry and target points\cite{straighttrack}, adding a further improvement by showing an arrow in the direction that the surgeon needs to adjust their approach to align to the planned trajectory. Furthermore, we track the tip position and print the current depth to target which we compute by projecting the tool trajectory onto the planned trajectory and then computing the distance to the target point. 

\begin{table}[tb]
  \centering
  \caption{Cohort Characteristics}
  \label{tab:cohort}
  \scriptsize
  \setlength{\tabcolsep}{3pt}%
  \renewcommand{\arraystretch}{1.0}%
  \begin{tabular}{l r}
    \toprule
    \textbf{Parameter} & \textbf{Value (n)} \\
    \midrule
    \textbf{Sex} & \\
    \quad Male & 7 \\
    \quad Female & 2 \\
    \addlinespace[5pt]
    \textbf{Handedness} & \\
    \quad Right & 9 \\
    \addlinespace[5pt]
    \textbf{Training Level} & \\
    \quad Engineer & 2 \\
    \quad Junior Neurosurgery Resident& 3 \\
    \quad Senior Neurosurgery Resident& 2 \\
    \quad Neurosurgery Faculty & 2 \\
    \addlinespace[5pt]
    \textbf{Experience with conventional navigation} & \\
    \quad None & 2 \\
    \quad 3--5 sessions & 1 \\
    \quad 10+ sessions & 6 \\
    \addlinespace[5pt]
    \textbf{Experience with Augmented Reality navigation} & \\
    \quad None & 3 \\
    \quad 1--2 sessions & 1 \\
    \quad 3--5 sessions & 0 \\
    \quad 5--10 sessions & 2 \\
    \quad 10+ sessions & 3 \\
    \bottomrule
  \end{tabular}
\end{table}

\subsection{User Study}
We recruited 9 users (2 neurosurgery faculty, 2 senior neurosurgical residents, 3 junior neurosugical residents, 2 engineers). Each user provided informed consent as approved by our institutional review board and performed 12 simulated EVD placements on 3D-printed head phantoms in a balanced within-subject design (6 trials for in-situ vs.\ 6 for Tool-Tracking), following randomized planned trajectories \begin{table}[tb]
  \centering
  \caption{Overview of Experimental Design}
  \label{tab:study_design}
  \scriptsize
  \begin{tabular}{p{0.37\linewidth} p{0.55\linewidth}}
    \toprule
    \textbf{Independent Variables} & \\[-0.3ex]
    \midrule
    Users & 9 (varied background) \\
    Conditions   & 2 (in-situ vs.\ tool-tracking) \\
    Planned Trajectories & 12  \\
    \midrule
    \textbf{Dependent Variables} & \\[-0.3ex]
    \midrule
    Catheter Entry Offset (mm)    & Distance at the inner burr-hole / skull interface \\
    Angular Deviation ($^\circ$)         & Angle between planned and actual trajectories \\
    Target Depth Error (mm)       & Difference between catheter tip depth and plan \\
    Target Radial Error (mm)  & Lateral deviation of catheter tip at target \\
    Marking Time (s)              & Time to localize and mark each burr-hole entry \\
    Insertion Time (s)            & Time to advance catheter to planned depth \\
    Subjective Metrics            & System Usability Scale (SUS), 7-item Likert \\
    \bottomrule
  \end{tabular}
\end{table}

\subsubsection{Protocol Randomization}
\label{sec:protocol_randomization}

We used a custom Python script to create randomized, balanced protocols for each user. The script first shuffled a set of predefined left- and right-sided trajectories, then assigned them to two visualization conditions (\emph{in-situ} and \emph{tool-tracking}) and two procedural phases (\emph{Marking} and \emph{Catheter Insertion}). By ensuring every user encountered each condition in a randomized order while preserving balance across sides and conditions, the protocol minimized learning effects and potential biases. The generated task list for each user was logged in a standardized format that included the action state (e.g., marking, catheter insertion), visualization mode (in-situ, tool-tracking), side (left or right), and a unique trajectory key.

\subsubsection{Equipment Setup}
We printed 20 phantom models from polylactic acid using a BambuLab A1 printer based on patient CT data for which we had consent to use in this study (dimensions: 223 mm $\times$ 210 mm $\times$ 114 mm). The heads were filled with ballistic gel to simulate brain tissue following Van Gestel et al.\ \cite{VanGestel2021_AR_EVD}. We purchased 1.5 mm wide, 150 mm long metal rods to serve as simulated catheters which were verified with a caliper. A replica of a commercially-available passive catheter introducer (PCI) instrument was produced from polylactic acid for insertion of the simulated catheter into the phantom, and was designed to be easily detachable. 
For each user, identical trajectories were planned to guide drilling and insertion of the catheter. We selected entry points for drilling and target points in the ventricles with two surgeon collaborators. The insertion point for the catheter was determined along the trajectory from the entry point on the skin to the target point on the bone.

\subsubsection{Experiment Workflow}
Surgeons conducted an eye calibration with the HoloLens 2, then performed one practice registration and three practice drill and insertion tests following prompts in the AR application. After practice on a different head phantom, landmark-based registration was performed followed by a surface tracing refinement. Trajectories were displayed using the UI described in \ref{subsubsec:guidance}. They were randomized and balanced, and the user simply had to say ``next trajectory'' to go through all of them.
For each trajectory, first guidance was provided to find the entry point on skin for drilling, which were marked by the surgeon with a pen. Then they drilled the holes. The models were all rigidly attached to a table at a comfortable height for the user via screwing them into a wooden plate and clamping the plate to the table. After drilling all the holes, guidance was provided for the PCI placement. After all simulated catheters were placed using the PCI array, they were glued into place to prevent any movement during transport to the CT scanner.
\begin{figure*}[t]
  \centering

\begin{subfigure}[t]{0.23\textwidth}
  \centering
  \includegraphics[width=\linewidth]{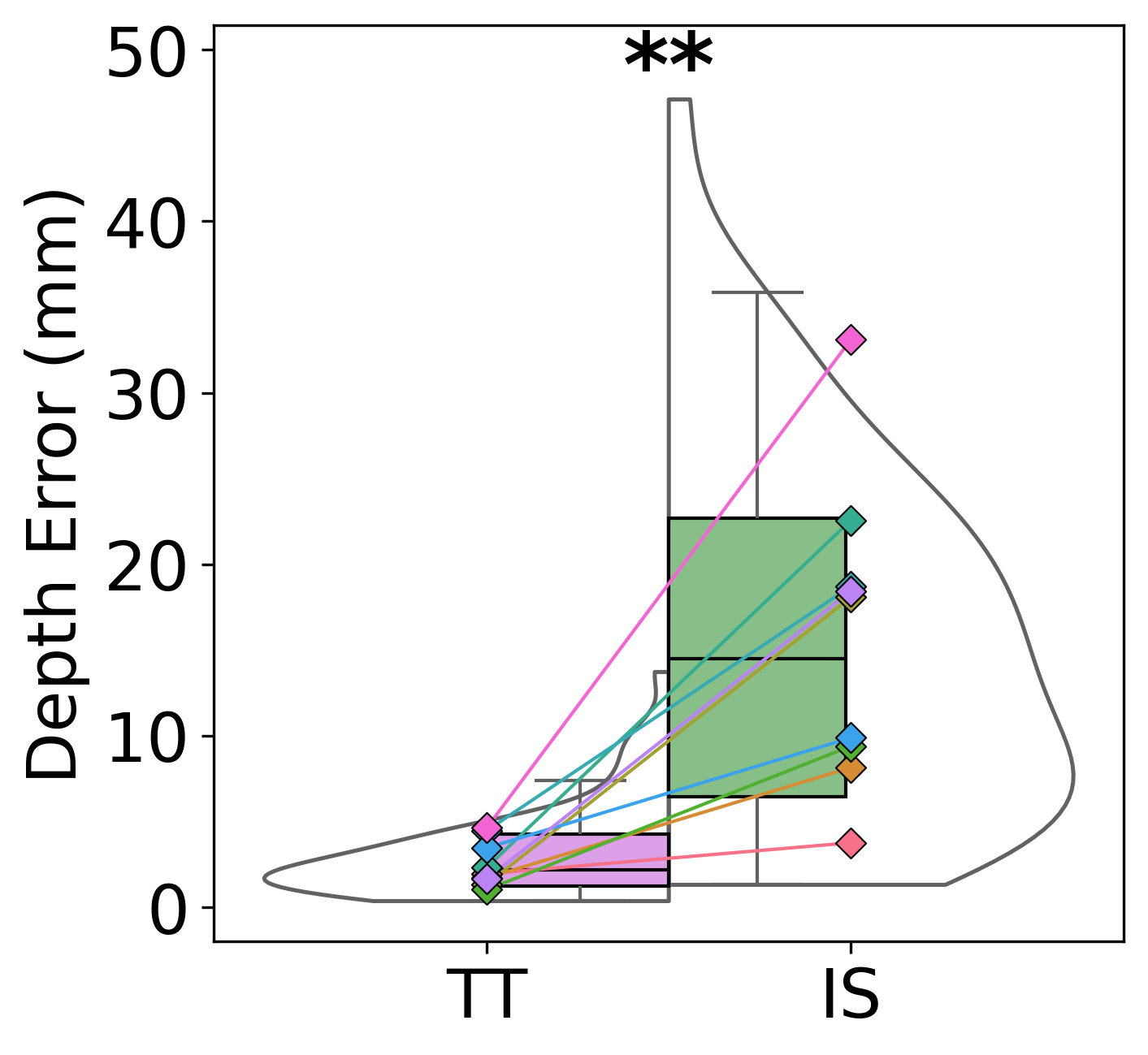}\\[0ex]
  \centering
\resizebox{\linewidth}{!}{%
\scriptsize
\setlength{\tabcolsep}{2pt}
\renewcommand{\arraystretch}{1.25}
\begin{tabular}{|>{\centering\arraybackslash}m{0.22\textwidth}|>{\centering\arraybackslash}m{0.22\textwidth}|>{\centering\arraybackslash}m{0.22\textwidth}|>{\centering\arraybackslash}m{0.22\textwidth}|>{\centering\arraybackslash}m{0.22\textwidth}|}
\hline
M TT & IQR TT & M IS & IQR IS & p-value \\
\hline
2.2 & 3.0 & 14.5 & 16.2 & 0.00391 \\
\hline
\end{tabular}%
}
  \caption{Target Depth Error (mm)}
  \label{subfig:pooled_depth_err}
\end{subfigure}\hspace{0.01\textwidth}
\begin{subfigure}[t]{0.23\textwidth}
  \centering
  \includegraphics[width=\linewidth]{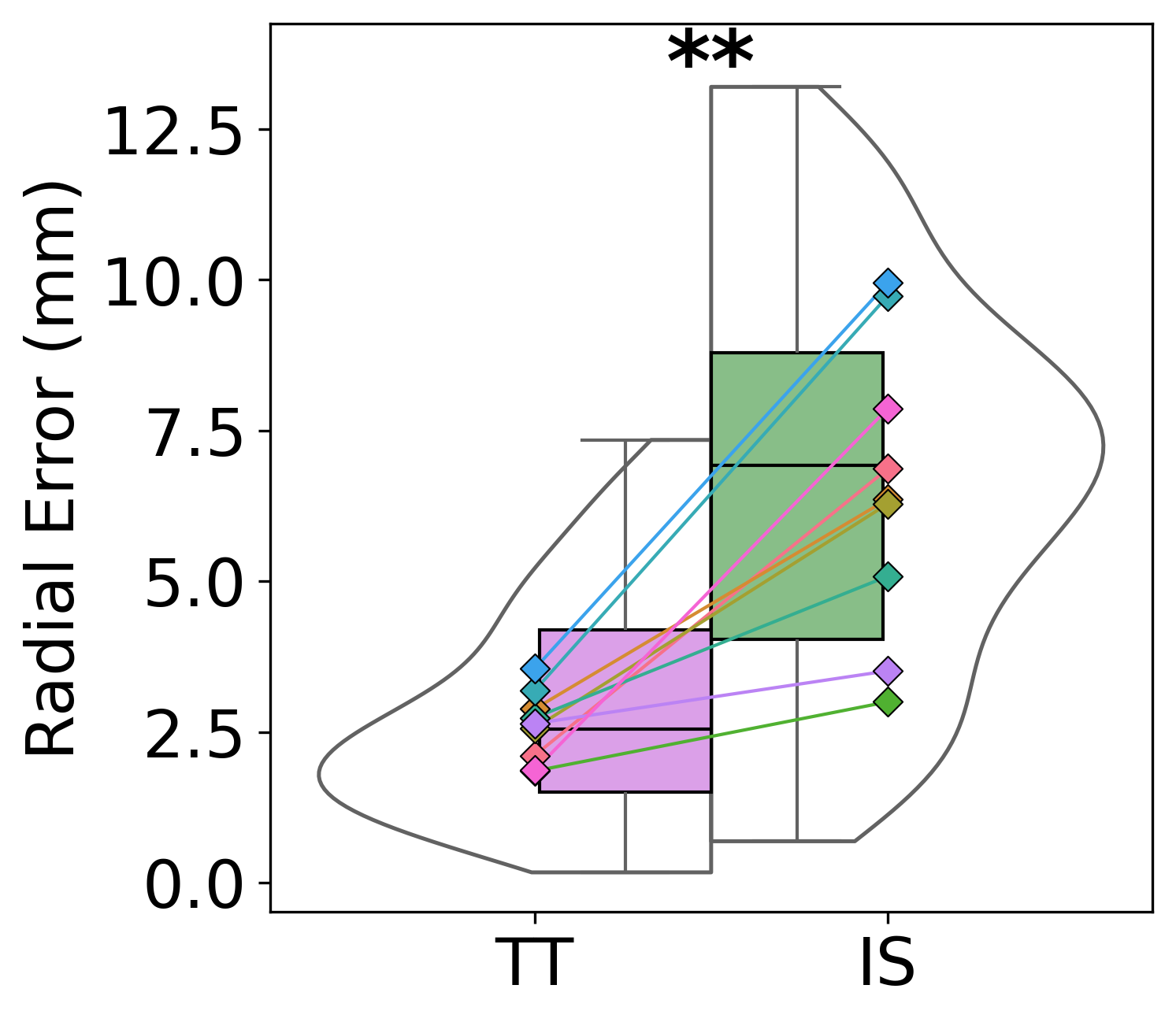}\\[0ex]
  \centering
\resizebox{\linewidth}{!}{%
\scriptsize
\setlength{\tabcolsep}{2pt}
\renewcommand{\arraystretch}{1.25}
\begin{tabular}{|>{\centering\arraybackslash}m{0.22\textwidth}|>{\centering\arraybackslash}m{0.22\textwidth}|>{\centering\arraybackslash}m{0.22\textwidth}|>{\centering\arraybackslash}m{0.22\textwidth}|>{\centering\arraybackslash}m{0.22\textwidth}|}
\hline
M TT & IQR TT & M IS & IQR IS & p-value \\
\hline
2.5 & 2.7 & 6.9 & 4.7 & 0.00391 \\
\hline
\end{tabular}%
}
  \caption{Target Radial Error (mm)}
  \label{subfig:pooled_orth_err}
\end{subfigure}%
\begin{subfigure}[t]{0.23\textwidth}
  \centering
  \includegraphics[width=\linewidth]{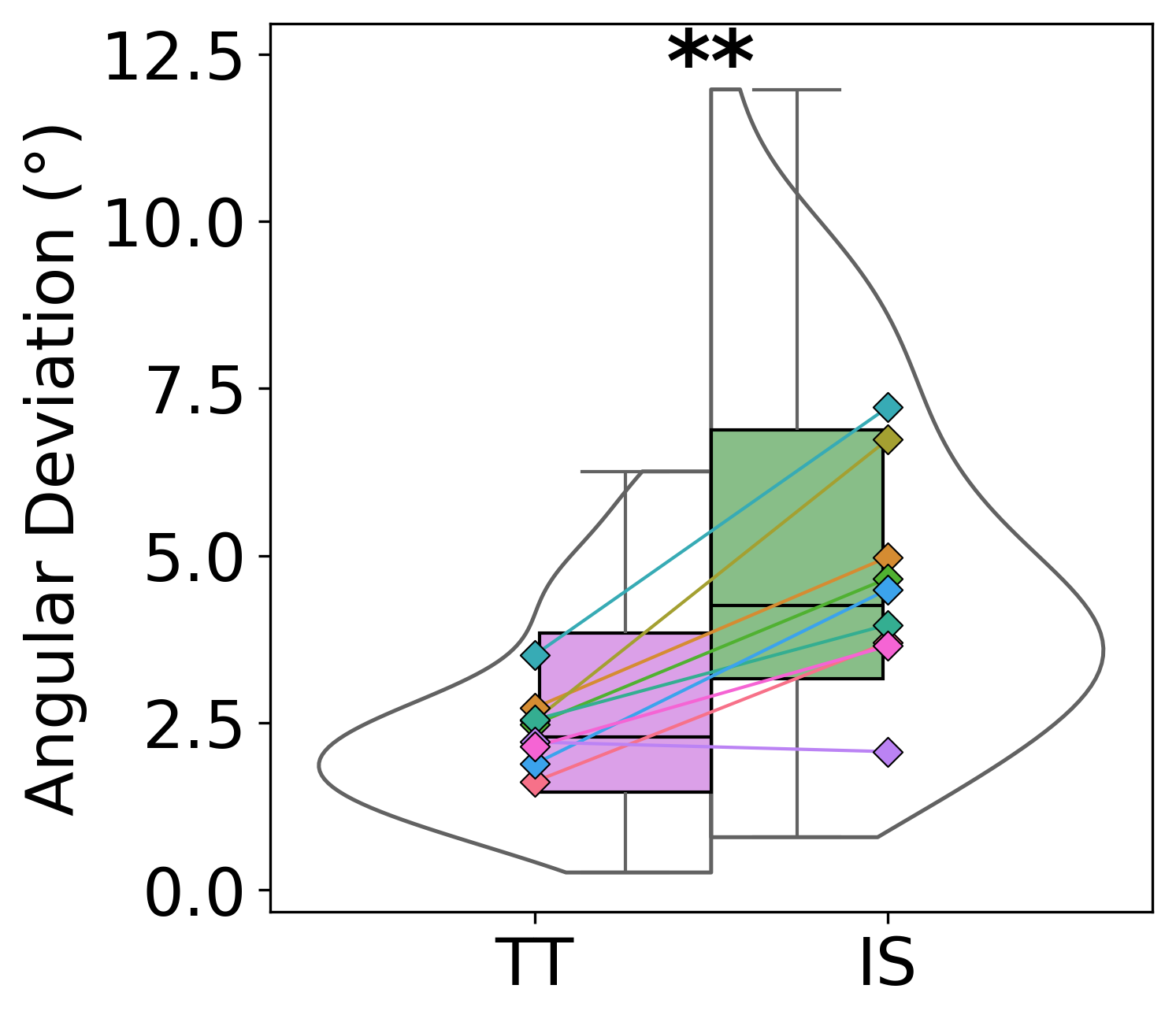}\\[0ex]
  \centering
\resizebox{\linewidth}{!}{%
\scriptsize
\setlength{\tabcolsep}{2pt}
\renewcommand{\arraystretch}{1.25}
\begin{tabular}{|>{\centering\arraybackslash}m{0.22\textwidth}|>{\centering\arraybackslash}m{0.22\textwidth}|>{\centering\arraybackslash}m{0.22\textwidth}|>{\centering\arraybackslash}m{0.22\textwidth}|>{\centering\arraybackslash}m{0.22\textwidth}|}
\hline
M TT & IQR TT & M IS & IQR IS & p-value \\
\hline
2.3 & 2.4 & 4.3 & 3.7 & 0.00781 \\
\hline
\end{tabular}%
}
  \caption{Angular Deviation ($^\circ$)}
  \label{subfig:pooled_angle_dev}
\end{subfigure}\hspace{0.01\textwidth}
\begin{subfigure}[t]{0.23\textwidth}
  \centering
  \includegraphics[width=\linewidth]{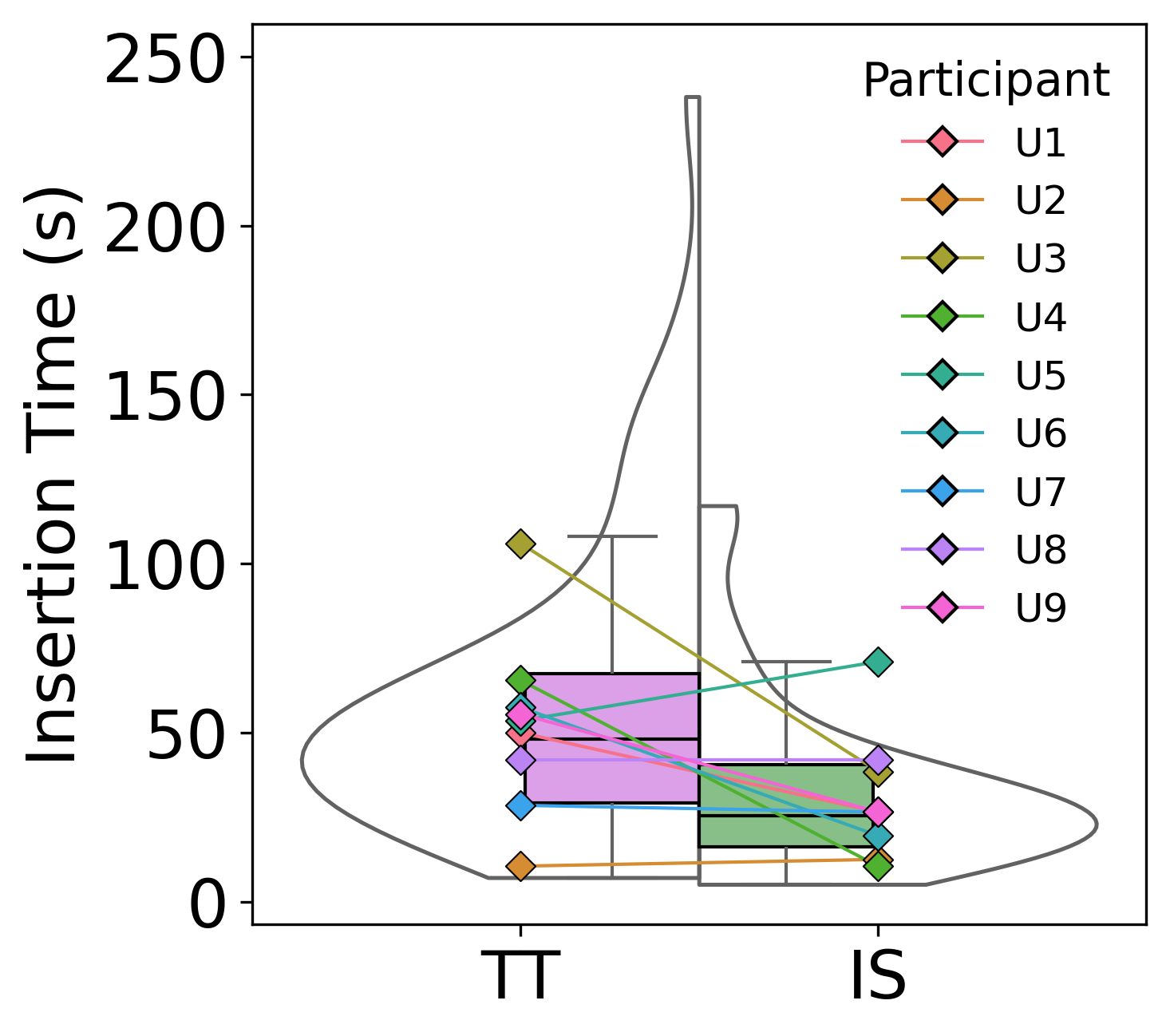}\\[0ex]
  \centering
\resizebox{\linewidth}{!}{%
\scriptsize
\setlength{\tabcolsep}{2pt}
\renewcommand{\arraystretch}{1.25}
\begin{tabular}{|>{\centering\arraybackslash}m{0.22\textwidth}|>{\centering\arraybackslash}m{0.22\textwidth}|>{\centering\arraybackslash}m{0.22\textwidth}|>{\centering\arraybackslash}m{0.22\textwidth}|>{\centering\arraybackslash}m{0.22\textwidth}|}
\hline
M TT & IQR TT & M IS & IQR IS & p-value \\
\hline
48.0 & 38.2 & 25.5 & 24.2 & 0.0584 \\
\hline
\end{tabular}%
}
  \caption{Insertion Time (s)}
  \label{subfig:pooled_insert_time}
\end{subfigure}%

\begin{subfigure}[t]{0.23\textwidth}
  \centering
  \includegraphics[width=\linewidth]{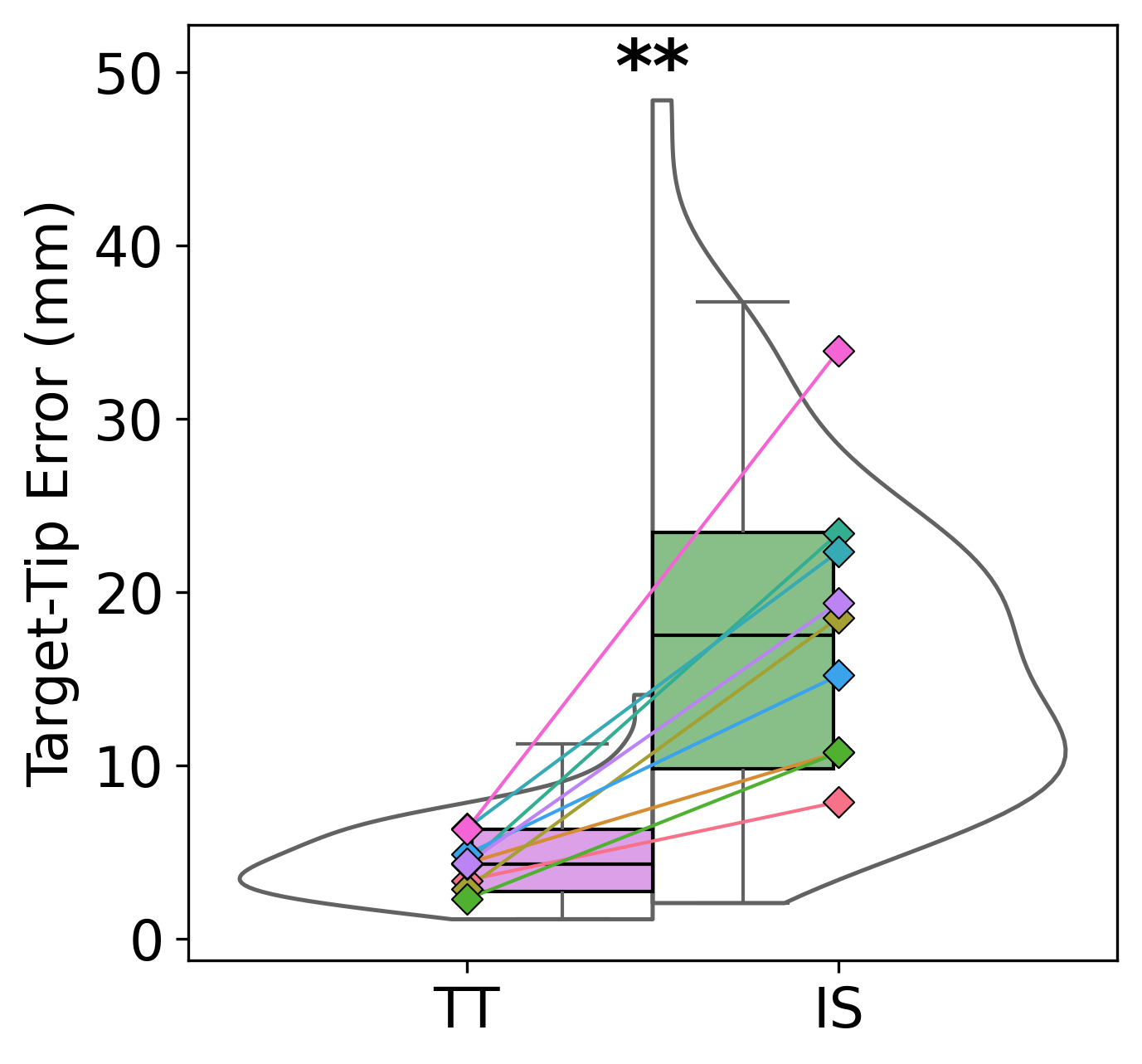}\\[0ex]
  \centering
\resizebox{\linewidth}{!}{%
\scriptsize
\setlength{\tabcolsep}{2pt}
\renewcommand{\arraystretch}{1.25}
\begin{tabular}{|>{\centering\arraybackslash}m{0.22\textwidth}|>{\centering\arraybackslash}m{0.22\textwidth}|>{\centering\arraybackslash}m{0.22\textwidth}|>{\centering\arraybackslash}m{0.22\textwidth}|>{\centering\arraybackslash}m{0.22\textwidth}|}
\hline
M TT & IQR TT & M IS & IQR IS & p-value \\
\hline
4.3 & 3.6 & 17.5 & 13.6 & 0.00391 \\
\hline
\end{tabular}%
}
  \caption{Target-Tip Error (mm)}
  \label{subfig:pooled_target_tip_err}
\end{subfigure}\hspace{0.01\textwidth}
\begin{subfigure}[t]{0.23\textwidth}
  \centering
  \includegraphics[width=\linewidth]{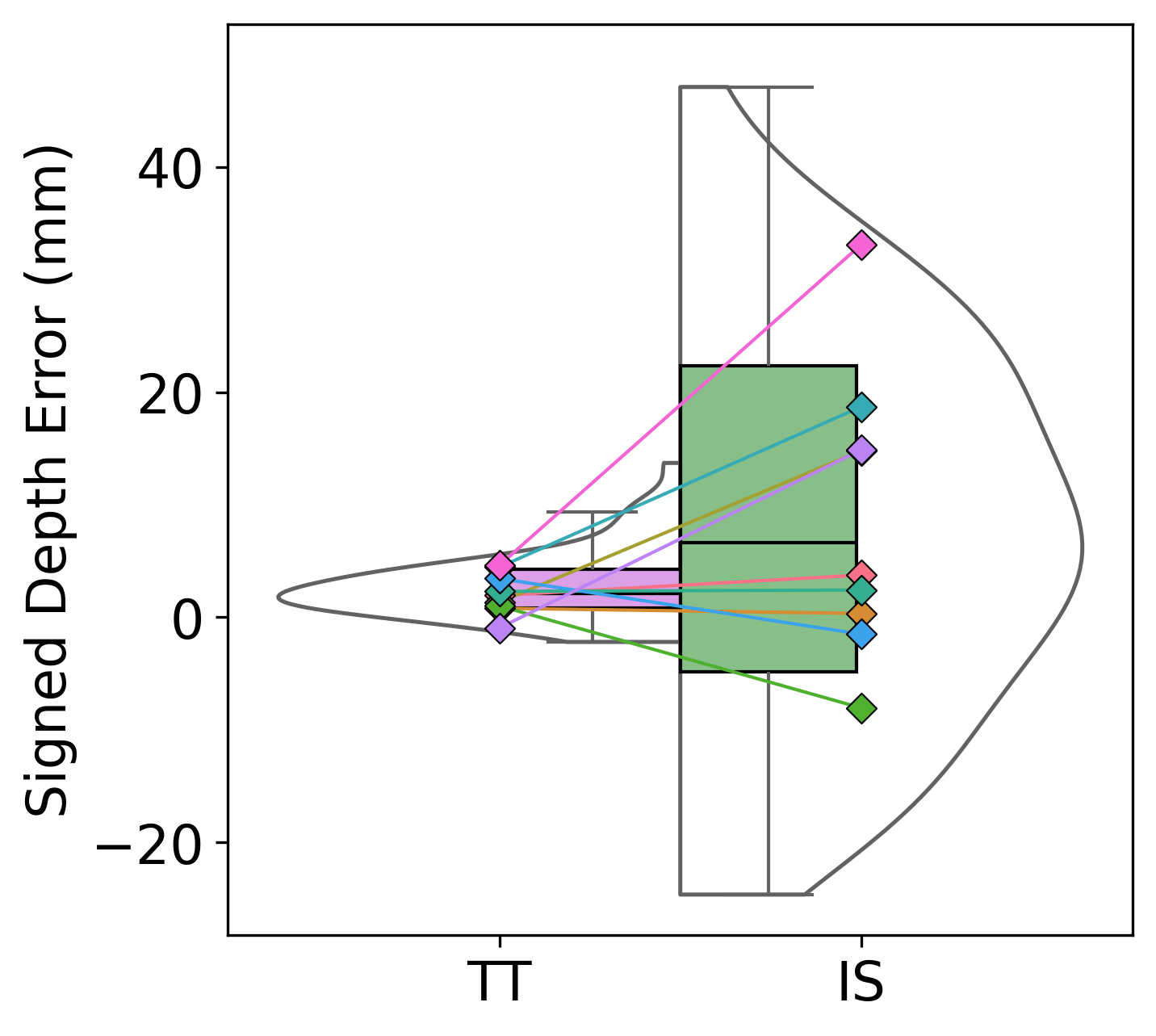}\\[0ex]
  \centering
\resizebox{\linewidth}{!}{%
\scriptsize
\setlength{\tabcolsep}{2pt}
\renewcommand{\arraystretch}{1.25}
\begin{tabular}{|>{\centering\arraybackslash}m{0.22\textwidth}|>{\centering\arraybackslash}m{0.22\textwidth}|>{\centering\arraybackslash}m{0.22\textwidth}|>{\centering\arraybackslash}m{0.22\textwidth}|>{\centering\arraybackslash}m{0.22\textwidth}|}
\hline
M TT & IQR TT & M IS & IQR IS & p-value \\
\hline
2.1 & 3.4 & 6.7 & 27.2 & 0.203 \\
\hline
\end{tabular}%
}
  \caption{Signed Depth Error (mm)}
  \label{subfig:pooled_signed_depth_err}
\end{subfigure}\hspace{0.01\textwidth}
\begin{subfigure}[t]{0.23\textwidth}
  \centering
  \includegraphics[width=\linewidth]{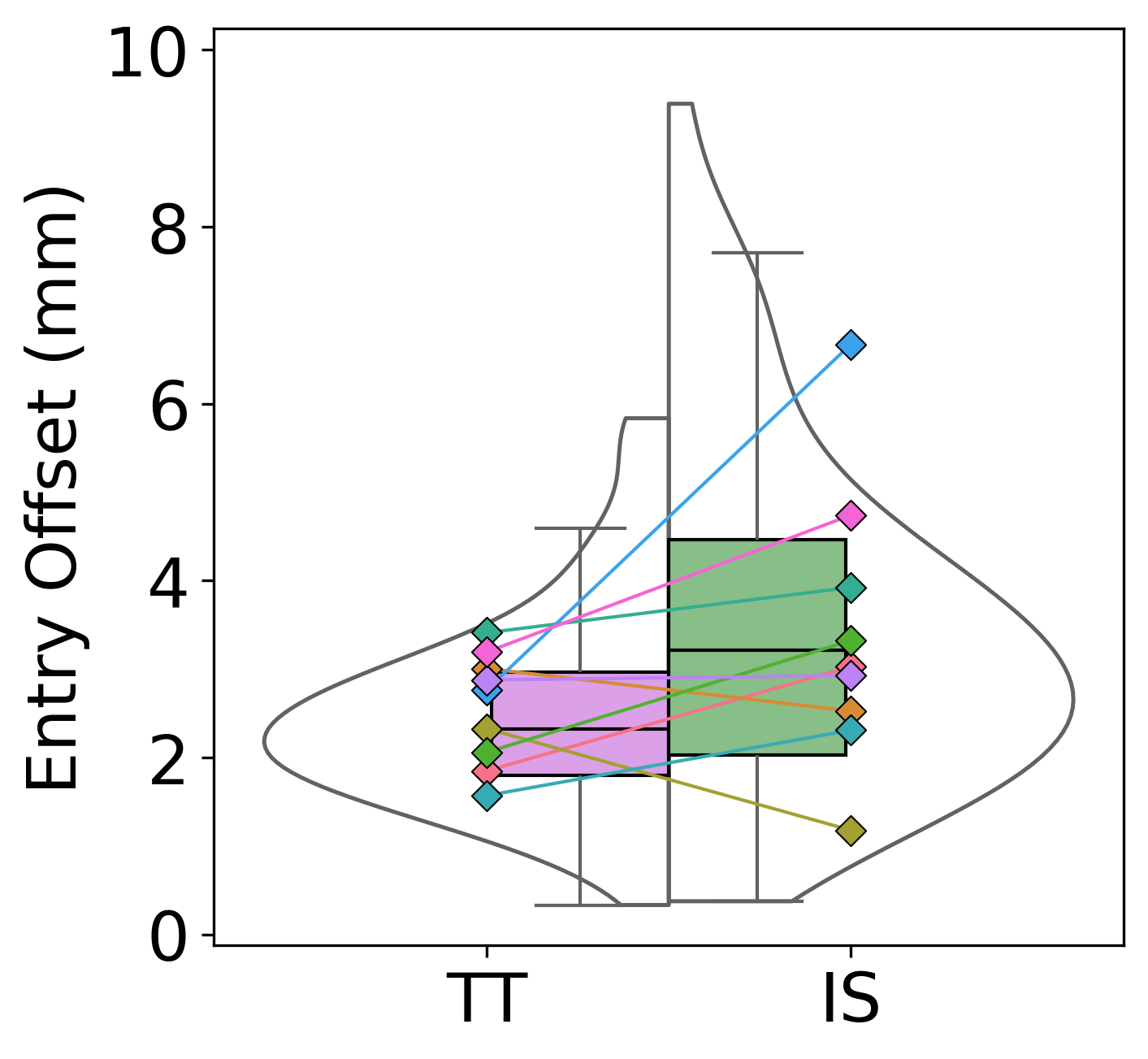}\\[0ex]
  \centering
\resizebox{\linewidth}{!}{%
\scriptsize
\setlength{\tabcolsep}{2pt}
\renewcommand{\arraystretch}{1.25}
\begin{tabular}{|>{\centering\arraybackslash}m{0.22\textwidth}|>{\centering\arraybackslash}m{0.22\textwidth}|>{\centering\arraybackslash}m{0.22\textwidth}|>{\centering\arraybackslash}m{0.22\textwidth}|>{\centering\arraybackslash}m{0.22\textwidth}|}
\hline
M TT & IQR TT & M IS & IQR IS & p-value \\
\hline
2.3 & 1.2 & 3.2 & 2.4 & 0.0742 \\
\hline
\end{tabular}%
}
  \caption{Catheter Entry Offset (mm)}
  \label{subfig:pooled_entry_offset}
\end{subfigure}\hspace{0.01\textwidth}
\begin{subfigure}[t]{0.23\textwidth}
  \centering
  \includegraphics[width=\linewidth]{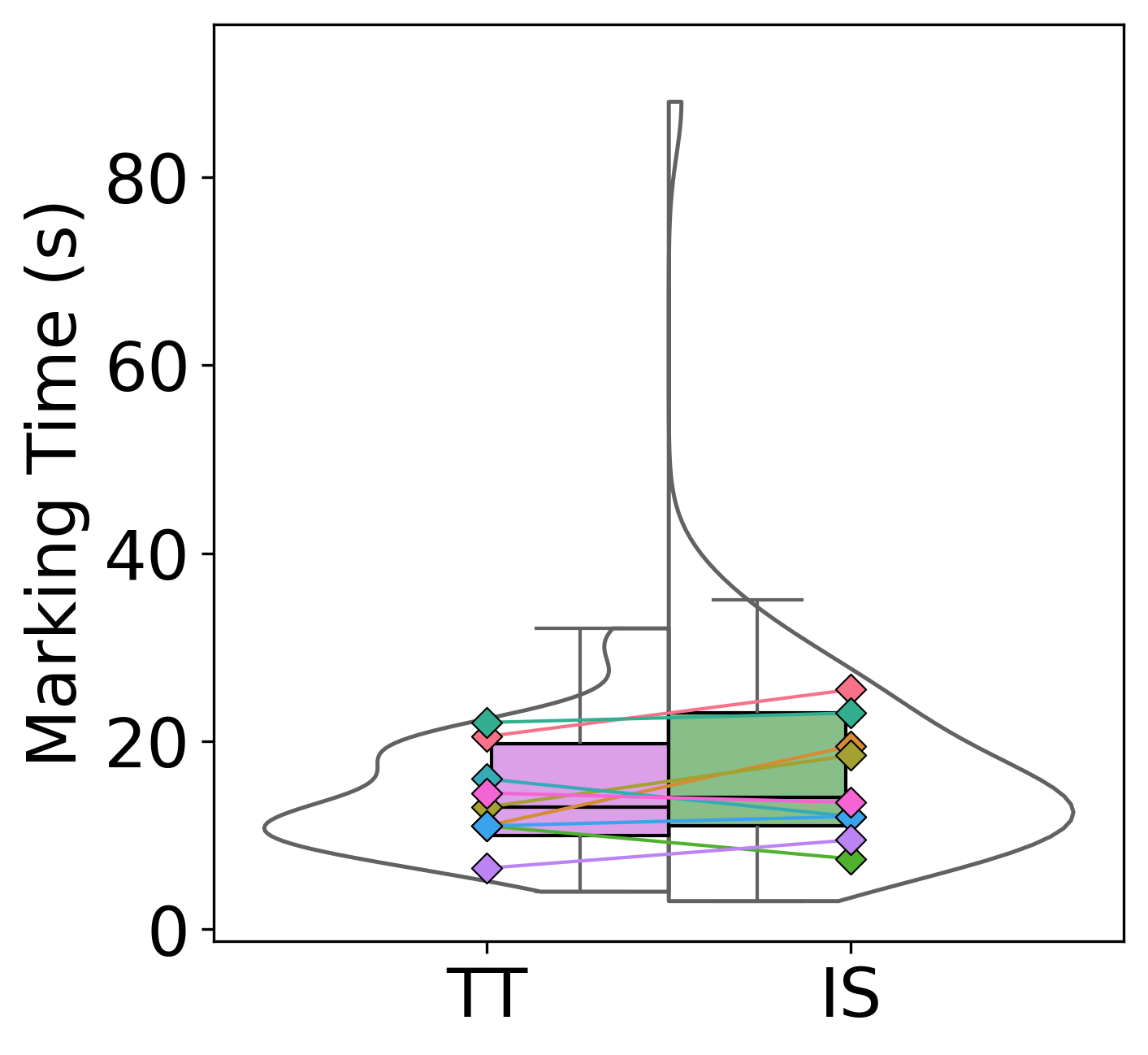}\\[0ex]
  \centering
\resizebox{\linewidth}{!}{%
\scriptsize
\setlength{\tabcolsep}{2pt}
\renewcommand{\arraystretch}{1.25}
\begin{tabular}{|>{\centering\arraybackslash}m{0.22\textwidth}|>{\centering\arraybackslash}m{0.22\textwidth}|>{\centering\arraybackslash}m{0.22\textwidth}|>{\centering\arraybackslash}m{0.22\textwidth}|>{\centering\arraybackslash}m{0.22\textwidth}|}
\hline
M TT & IQR TT & M IS & IQR IS & p-value \\
\hline
13.0 & 9.8 & 14.0 & 12.0 & 0.301 \\
\hline
\end{tabular}%
}
  \caption{Marking Time (s)}
  \label{subfig:pooled_mark_time}
\end{subfigure}\hspace{0.01\textwidth}

  \caption{Pooled-user performance under tool-tracking (TT) and in-situ (IS) visualisation across eight metrics.  
(a) \textbf{Target-depth error}: absolute difference between planned and achieved depth along the trajectory.  
(b) \textbf{Target-radial error}: lateral (orthogonal) offset of the catheter tip from the planned target.  
(c) \textbf{Angular deviation}: angle between the planned and actual catheter trajectories.  
(d) \textbf{Insertion time}: elapsed time from catheter entry until the planned depth is reached.  
(e) \textbf{Target--tip error}: 3-D Euclidean distance between the catheter tip and the target point.  
(f) \textbf{Signed depth error}: signed over-/under-shoot relative to the planned depth (positive = too deep).  
(g) \textbf{Catheter-entry offset}: 3-D Euclidean distance between planned and actual entry points on the brain surface.  
(h) \textbf{Marking time}: time needed to localise and mark each burr-hole entry site.  
Tables within each panel list the median (M), inter-quartile range (IQR), and Wilcoxon signed-rank p-value comparing TT with IS for that metric.}%

  \label{fig:pooled_metrics}
\end{figure*}

\begin{table*}[htbp]
  \centering
  \caption{Likert Questionnaire Comparison of In-Situ vs.\ Tool Tracking (n=9 each)}
  \label{tab:likert}
  \scriptsize
  \setlength{\tabcolsep}{4pt}%
  \renewcommand{\arraystretch}{1.0}%
  \resizebox{\textwidth}{!}{%
  \begin{tabular}{r l c c c c}
    \toprule
    \# & \textbf{Likert Questions} & 
    \textbf{In-Situ} & 
    \textbf{Tool Tracking} & 
    \textbf{Difference} & 
    \textbf{P-Value}\\
    \midrule
    1 & Intuitive to use 
      & 3.7 (1.06) & 4.4 (0.96) & -0.07 & 0.140 \\
    2 & Easy to mark the drilling point 
      & 3.4 (0.96) & 4.6 (0.52) & -1.2 & 0.003 \\
    3 & Easy to visualize angulation of the instrument prior to insertion 
      & 2.8 (1.3) & 4.2 (0.79) & -1.4 & 0.010 \\
    4 & Easy to visualize depth to target  
      & 1.78 (1.32) & 4.7 (0.67) & -2.92 & $<.001$ \\
    5 & Comparable to commercially-available neuro-navigation \newline
      & 2.4 (1.17) & 4.3 (0.71) & -1.9 & $<.001$ \\
    6 & More confidence hitting target vs.\ freehand \newline
        (e.g., freehand EVD placement) 
      & 3.4 (0.84) & 4.6 (0.69) & -1.2 & 0.003 \\
    7 & Provides additional value vs.\ conventional surgical navigation 
      & 3.1 (1.197) & 4.4 (0.84) & -1.3 & 0.012 \\
    \bottomrule
  \end{tabular}
  }
\end{table*}

\begin{table*}[htbp]
\centering
\begin{tabular}{p{0.7\textwidth}c}
\toprule
Question & Mean $\pm$ SD \\
\midrule
I think that I would like to use this system frequently & 4.6 $\pm$ 0.7 \\
I found the system unnecessarily complex & 1.9 $\pm$ 0.57 \\
I thought the system was easy to use & 4.3 $\pm$ 0.67 \\
I think that I would need the support of a technical person to be able to use this system & 2.5 $\pm$ 1.08 \\
I found the various functions in this system were well integrated & 4.4 $\pm$ 0.52 \\
I thought there was too much inconsistency in this system & 2.1 $\pm$ 0.88 \\
I would imagine that most people would learn to use this system very quickly & 4.3 $\pm$ 0.67 \\
I found the system very cumbersome to use & 2.0 $\pm$ 0.67 \\
I felt very confident using the system & 4.1 $\pm$ 0.88 \\
I needed to learn a lot of things before I could get going with this system & 1.8 $\pm$ 0.92 \\
\bottomrule
\end{tabular}
\caption{System Usability Scale (SUS) item results (Likert 1--5). Overall SUS $= 78.5 \pm 13.2$ ($n=9$).}
\label{tab:sus_results}
\end{table*}

\subsubsection{Validation Assessment}
After all EVDs were placed, the phantom underwent a post-operative CT scan. All 3D printed head phantoms were scanned on the Aquilion One Insight Edition CT scanner (Canon Medical Systems USA) at 120 kV using the volume scan mode with a CTDI vol,16 value of 57.7 mGy. The acquisition data was reconstructed using a deep learning technique (PIQE) with a body kernel and a voxel size of 0.252 mm $\times$ 0.252 mm $\times$ 0.500 mm. For the CT acquisitions after needle placement, a metal artifact reduction technique (SEMAR) was employed to minimize artifacts caused by the metal material of the metal rods. These images were registered to the pre-operative CT using 7 landmarks (right and left tragal notch, lateral and medial canthi and subansale) and ICP refinement with mean error. We fit a line to segmented metal trajectories and calculated the intersection point with the skin surface outside and inside and the end point. We then compared the outside intersection with the planned skin entry point, the inside intersection with the planned catheter entry point and the end point with the planned target point. The translational error (distance between planned and actual entry/target points) and angular error (angular difference from the planned trajectories) were computed. Procedure duration was recorded for each trajectory in each visualization group, as well as for the landmark based registration and the surface tracing. Surveys were completed for qualitative assessment. Statistical analyses were performed to compare accuracy metrics, workflow efficiency, and user preference between the in-situ and tool-tracked guidance.

\section{Statistical Analysis}

The study employed a within-subjects design with 9 users each performing 12 simulated EVD placement trials (6 trials per visualization technique) along predefined trajectories into a phantom. We used a CT of the phantom to pre-plan the 12 trajectories (6 left, 6 right).  
For each user these 12 trajectories were carried out twice---first for drill-hole marking, then for rod insertion---in the same sequence.  
Visualization modes were balanced (3 left + 3 right per mode), and both the trajectory-mode pairing and the execution order were independently randomised for every user, yielding exactly six \textit{tool-tracking} and six \textit{in-situ} trials.

For each trial, we recorded continuous dependent variables including entry point error (mm), depth error, radial error and orientation error (degrees) between the placed rod and the planned trajectory. We also recorded the time it took users to do each marking and each insertion. Data distribution was assessed through visual inspection of histograms and violin plots, as well as formal testing using the Shapiro--Wilk test. This analysis revealed non-normal distributions.

\medskip
\noindent
Given the non-normal data distribution, within-subjects design, and comparison between two conditions tool-tracking and in-situ, we employed the Wilcoxon signed-rank test for statistical comparison. To satisfy the test's independence requirement, the six trial values for each surgeon $\times$ condition were collapsed to a single median, yielding one tool-tracking in-situ pair per user. All outcomes are continuous and on identical scales, and inspection of difference distributions revealed no marked skew, meeting the test's remaining assumptions.  

\noindent
Statistical significance was set at $p < 0.05$. All statistical analyses were performed using \texttt{scipy.stats} in Python\,\cite{2020SciPy-NMeth}.

\section{Results}
\label{sec:results}

\subsection{Placement Accuracy}
\label{sec:placementaccuracy}

\Cref{fig:pooled_metrics} summarises pooled-user performance for each visualisation technique. Eight metrics were analysed: \emph{catheter entry offset}, \emph{angular deviation}, \emph{target--tip error}, \emph{target depth error}, \emph{signed depth error}, \emph{target radial error}, \emph{marking time}, and \emph{insertion time}. For each user the six trials per condition were collapsed to the median, yielding nine paired observations that were compared with the Wilcoxon signed-rank test at $\alpha = 0.05$.

\paragraph{Catheter Entry Offset.}
Tool-tracking guidance produced a median offset of $2.3$\,mm (IQR\,$1.2$\,mm) versus $3.2$\,mm (IQR\,$2.4$\,mm) for in-situ overlays; the difference did not reach statistical significance ($p = 0.0742$).

\paragraph{Angular Deviation.}
Median angular deviation fell from $4.3^\circ$ (IQR\,$3.7^\circ$) with in-situ to $2.3^\circ$ (IQR\,$2.4^\circ$) under tool-tracking ($p = 0.00781$), indicating that continuous feedback helped users maintain the planned trajectory.

\paragraph{Target--Tip Error.}
Tool-tracking reduced the final 3-D distance between catheter tip and planned target by roughly 75\%: median $4.3$\,mm (IQR\,$3.6$\,mm) compared with $17.5$\,mm (IQR\,$13.6$\,mm) for in-situ ($p = 0.00391$).

\paragraph{Target Depth Error.}
Over-/undershoot was also greatly diminished, from $14.5$\,mm (IQR\,$16.2$\,mm) with in-situ to $2.2$\,mm (IQR\,$3.0$\,mm) with tool-tracking ($p = 0.00391$).

\paragraph{Signed Depth Error.}
Although tool-tracking showed a lower median signed depth error ($2.1$\,mm, IQR\,$3.4$\,mm) than in-situ ($6.7$\,mm, IQR\,$27.2$\,mm), the difference was not significant ($p = 0.203$), reflecting large variability in the direction of depth deviations.

\paragraph{Target Radial Error.}
Lateral deviation at the intracranial target dropped from $6.9$\,mm (IQR\,$4.7$\,mm) with in-situ to $2.5$\,mm (IQR\,$2.7$\,mm) with tool-tracking ($p = 0.00391$).

\paragraph{Marking and Insertion Times.}
Median marking times were similar---$13.0$\,s (IQR\,$9.8$\,s) for tool-tracking versus $14.0$\,s (IQR\,$12.0$\,s) for in-situ ($p = 0.301$). Insertion took longer with tool-tracking (median $48.0$\,s, IQR\,$38.2$\,s) than with in-situ ($25.5$\,s, IQR\,$24.2$\,s), but the difference remained non-significant ($p = 0.0584$).

\subsection{User Feedback and Workflow}
User-reported outcomes mirrored the objective accuracy gains. Likert-scale responses (\Cref{tab:likert}) show that tool-tracking was rated higher for identifying drilling points, gauging instrument angulation, and estimating depth. System Usability Scale scores again placed tool-tracking in the ``Good--Excellent'' range, while in-situ reached ``Good''. Landmark registration (median 47 s) followed by surface tracing (median 62 s) together required under two minutes, a duration participants deemed clinically acceptable.

\subsection{SUS Outcomes}
\label{sec:sus_outcomes}

To gauge the general usability of the \emph{entire} AR navigation workflow---
that is, landmark \& surface-tracing registration, user interface, and overall
interaction paradigm---participants completed the standard 10-item
\textbf{System Usability Scale (SUS)} questionnaire \emph{after} finishing all
trials.  The survey was framed to capture their impression of the system as a
whole; it did \emph{not} ask them to differentiate between the
\textit{tool-tracking} and \textit{in-situ} visualisation modes, which we
evaluated separately through the dedicated 7-item Likert instrument
(\cref{tab:likert}).

The aggregated SUS score was
\textbf{$78.5 \pm 13.2$} (mean $\pm$ SD, $n = 9$).  On the conventional
0 to 100 SUS scale, a score of~$\approx\!78$ lies in the
\emph{upper-``Good'' to lower-``Excellent''}
band---well above the widely cited acceptability threshold of 68---indicating that
users found the AR workflow both learnable and efficient even without
distinguishing between guidance modalities.

\section{Discussion}
\label{sec:discussion}

\subsection{Influence of Real-Time Tool Tracking on Accuracy}
Real-time tool tracking yielded pronounced accuracy gains across nearly every metric. Most notably, median target--tip error fell from 17.5\,mm to 4.3\,mm, and depth error from 14.5\,mm to 2.2\,mm, while lateral deviation and angular error were each roughly halved. Catheter entry offset improved modestly but not significantly, and signed depth error exhibited a non-significant trend in favour of tool-tracking. These results corroborate our hypothesis that continuous six-degree-of-freedom feedback mitigates both angular drift and excessive insertion, leading to more precise ventricular targeting. Much of the improvement stems from reduced angular error. Small deviations in trajectory quickly compound into large target offsets over the catheter's length. Tool tracking helps users correct these in real time, unlike static overlays.

\subsection{User Acceptance and Clinical Feasibility}
User responses indicate a strong preference for real-time feedback. Surgeons and residents alike found the tool-tracking interface more intuitive and felt it would reduce the risk of misplacement relative to both freehand and static AR overlays. This preference stems largely from the clear visualization of alignment and depth data as the catheter advances.  
Our surface-tracing registration workflow also demonstrated feasibility, although it added approximately one minute of setup time. In urgent cases (e.g., emergency EVD placement), minimizing registration time is crucial for clinical adoption, and future refinements should target further streamlining of this step.

\subsection{Limitations and Future Directions}
This study used 3D-printed head phantoms filled with ballistic gel rather than human tissue or cadavers. While phantoms ensure repeatable geometry and reduce confounding variables, they cannot fully replicate factors such as brain shift, variable tissue resistance, or the complexities of a live OR environment. Additionally, because the models lack a skin layer, excessive force during surface tracing could compress soft tissue in a real patient and introduce registration error. However, neurosurgeons are trained to apply minimal pressure, and our UI cues explicitly remind users to trace lightly. Moreover, although the CT-derived 3D models are geometrically one-to-one with the underlying bone, skin can swell or shift preoperatively, potentially degrading surface-based registration. To mitigate this, we perform surface-tracing registration on bony regions, which are far less susceptible to soft-tissue changes. Further trials on cadavers or in actual operative cases will be essential to validate performance under realistic conditions.

The sample size (n=9) and the range of experience levels could also limit generalizability. Nonetheless, the consistent trends across all experience strata suggest that even novice users benefit from tool-tracking feedback. Future work should explore adaptive guidance algorithms, integration with other intraoperative imaging modalities, and user interface optimizations to further enhance both accuracy and efficiency in clinical workflows.

\section{Conclusion}
\label{sec:conclusion}

We presented an augmented-reality neurosurgical navigation system that combines landmark- and surface-based registration with real-time tool tracking on a head-mounted display. In a controlled phantom study, the approach cut median target--tip error by more than fourfold (4.3\,mm vs 17.5\,mm) and reduced overshoot from 14.5\,mm to 2.2\,mm, while also halving angular and lateral deviations---all without materially lengthening the procedural workflow. Participants reported higher confidence and lower cognitive load under tool tracking compared with static in-situ visualisation. These findings underscore the promise of compact, easily deployed AR guidance for ventricular catheter placement and motivate continued evaluation in live clinical environments.
\acknowledgments{
We thank the UC Davis neurosurgical residents---Dr. Matthew Kercher, Dr. Jared Clouse, Dr. Jos\'e Castillo, and Dr. Khadija Soufi---for their invaluable assistance throughout the study.

Marc Fischer, Jeff Potts, Dax Jones, and  Brandon Strong are part of the Xironetic Founding Team and have a vested interest in the company.
}

\FloatBarrier
\bibliographystyle{abbrv-doi}

\bibliography{template}
\end{document}